\documentclass{article}

\usepackage[final, nonatbib]{neurips_2021}

\usepackage[utf8]{inputenc} 
\usepackage[T1]{fontenc} 
\usepackage{hyperref}
\usepackage{url}
\usepackage{booktabs}
\usepackage{amsfonts}
\usepackage{amsmath}
\usepackage{bm}
\usepackage{nicefrac}
\usepackage{microtype}
\usepackage{xcolor}
\usepackage{graphicx}
\usepackage{multirow}
\usepackage{wrapfig}
\usepackage{color}
\usepackage{float}
\usepackage{caption}
\definecolor{darkorchid}{rgb}{0.6, 0.2, 0.8}

\title{Reinforcement Explanation Learning}

\author{
  Siddhant Agarwal \thanks{agarwalsiddhant10@gmail.com} \quad
  Owais Iqbal \quad
  Sree Aditya Buridi \quad
  Madda Manjusha \quad
  Abir Das \\
  Indian Institute of Technology, Kharagpur\\
  India
}

\begin{document}

\maketitle
\vspace{-10pt}

\begin{abstract}
Deep Learning has become overly complicated and has enjoyed stellar success in solving several classical problems like image classification, object detection, etc. Several methods for explaining these decisions have been proposed. Black-box methods to generate saliency maps are particularly interesting due to the fact that they do not utilize the internals of the model to explain the decision. Most black-box methods perturb the input and observe the changes in the output. We formulate saliency map generation as a sequential search problem and leverage upon Reinforcement Learning (RL) to accumulate evidence from input images that most strongly support decisions made by a classifier. Such a strategy encourages to search \emph{intelligently} for the perturbations that will lead to high-quality explanations. While successful black box explanation approaches need to rely on heavy computations and suffer from small sample approximation, the deterministic policy learned by our method makes it a lot more efficient during the inference. Experiments on three benchmark datasets demonstrate the superiority of the proposed approach in inference time over state-of-the-arts without hurting the performance. Project Page: \texttt{\href{https://cvir.github.io/projects/rexl.html}{https://cvir.github.io/projects/rexl.html}}
\end{abstract}

\section{Introduction}

\label{Introduction}

Deep Learning methods are enjoying enormous success in computer vision, natural language and robotics research.
Over the years, the models have become overly complex and large with millions of parameters.
For example, GPT-$3$ \cite{gpt3} has about $175$ billion parameters.
Yet, it remains largely unclear how the system comes to a decision, how certain the model is about its decision, if and when it can be trusted or when it has to be corrected.
Explainable AI (XAI) is being extensively researched upon~\cite{Fong_2019_ICCV, Fong_2017_ICCV, Petsiuk2018RISE,  gradcam, springenberg2015, zeiler_fergus, excitation_backprop, cam}
to get a better insight into AI made decisions to not only improve the models but also to instill accountability in them.
The explanations could benefit users in many ways such as improving safety and fairness when relying on AI decisions, especially in safety critical applications \textit{e.g.}, healthcare \cite{health_care1, health_care2}, autonomous driving \cite{auto_drive1} or criminal justice \cite{compas}.

XAI Strategies for deep visual models offer model specific~\cite{neural_module_networks, lime}
or saliency map based explanations~\cite{Fong_2017_ICCV, Petsiuk2018RISE, norm, gradcam, cam}.
This work subscribes to the later where saliency maps highlight relevant regions in an input \textit{e.g.}, image, video or natural language texts.
Saliency maps are obtained either following a black box or a white/grey box technique.
White/grey box approaches \cite{gradcam, excitation_backprop, cam}
either assume access to the internals of the base model (the model whose predictions are explained) or modify the base model which can cost the base model its performance \cite{cam}.
This makes them unsuitable to be generalized over all architectures.

Black box methods~\cite{Fong_2019_ICCV, Fong_2017_ICCV, Petsiuk2018RISE, zeiler_fergus}
on the contrary, do not touch the internals of the base model.
Instead, the input image pixels are perturbed and passed through the base model.
The output prediction resulting in from the perturbed image differentiates between the important and non-important regions as perturbing important or relevant regions affects the output score more than the case when non-important image regions are perturbed.
A popular technique to achieve this is to view
saliency maps as weighted masks.
The perturbations to the input are performed by masking portions of the image using these weighted masks.
Different methods differ in the way these masks are constructed.
Fong \textit{et al.}~\cite{Fong_2019_ICCV, Fong_2017_ICCV} solve an optimization objective to construct optimal masks while RISE~\cite{Petsiuk2018RISE} uses large number of random masks for an image.
A good explanation is associated with a low deletion score~\cite{Petsiuk2018RISE} which measures the drop in probability of a class as pixels are gradually removed according to the saliency map given importance.

\begin{figure}[t!]
\centering
\includegraphics[width=140mm]{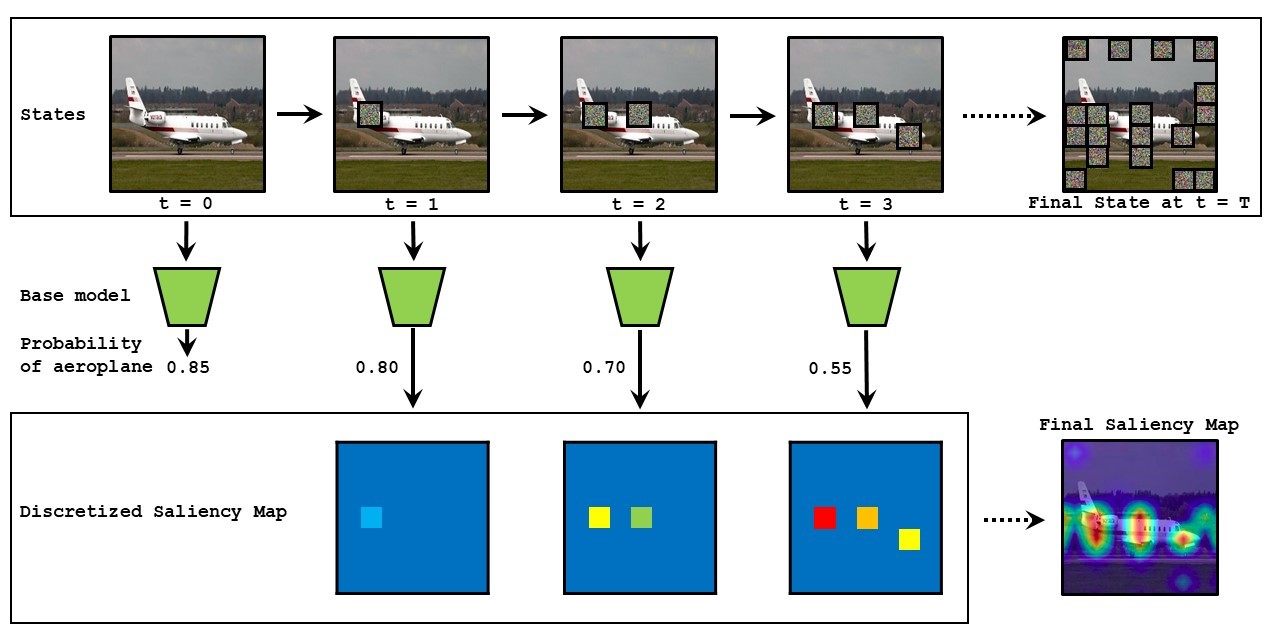}
\caption{\small \textbf{Approach Overview:} At the start, the base model (the classifier whose decision is explained) provides the probability of a class (here \emph{aeroplane}) in the unperturbed image.
Subsequently, at every step, the agent learns to choose important regions of the image to be masked such that on passing the masked image through the base model, the probability drops.
The drop in the probability gives the relevance of the masked region behind the base model's decision and is provided by the color-coded weight corresponding to the masked region in the saliency map (last row).
The weight value increases from blue to red.
Though initial importance of the chosen region at $t=1$ is not high, our approach provides due credit to it via a cumulating factor for driving the agent to choose regions leading to greater drop subsequently.
The final explanation is obtained by smoothening the discretized saliency map for better visualization.
(Best viewed in color.)
}
\label{fig:overview}
\end{figure}

In this paper, we address explainability as a sequential search problem where a sequence of image patches or regions are investigated in order to get the mask giving the saliency map.
Specifically, an agent starting with an unperturbed image, learns to choose the best image region to perturb given the choices made earlier.
The agent chooses the sequence of regions from the currently unperturbed portions of the image so that perturbing the region maximally drops the class probability which in turn, causes a low deletion score.
An approximation of the exhaustive search is performed by RISE~\cite{Petsiuk2018RISE} and is shown to achieve good deletion score albeit being prohibitively expensive both in time and memory requirements.
Our approach is fundamentally different as RISE does not make any conscious effort to get low deletion score whereas we \emph{learn to intelligently search} for the mask with a goal to achieve as low deletion score as quickly possible.
The goal oriented learning offered by the proposed method not only makes the inference fast but also makes it deterministic.

The core of the problem is to find a policy that assesses the image perturbed by the present mask and decides to perturb a new relevant image region.
Figure~(\ref{fig:overview}) illustrates the decision making process of such an agent.
At first, the unperturbed image is passed through the base classifier to get the probability of the class present in the image (an `aeroplane' in Figure~(\ref{fig:overview}))
The agent then starts by choosing a patch from the image and replacing it with uniform random noise.
The perturbed image is assessed by the base model to get the change in probability of the same class.
Subsequently, the agent decides to pick the most important region so that perturbing it decreases the probability of the class further until the agent decides to stop exploring or the maximum number of patches is perturbed.
As the perturbed regions contribute to a drop of class probability, these regions are pivotal evidences for the classifier to classify the image.
The significance of the perturbed regions behind the base classifier's decision is manifested by the drop in probability.
Thus at every step, the chosen region gets an weight proportional to the drop of the class probability.
This weighted regions provide a discrete weighted mask that after smoothening provides the saliency map explaining the decision of the classifier.
As region perturbation is sequential and its effect is cumulative on the base model, we introduce a cumulating factor $\lambda \in (0, 1)$ that partially credits the regions perturbed earlier but instead of immediate drop, may have resulted in a significant drop in class probability later.

Learning such a dynamic agent capable of performing sequential actions towards a long-term goal by actively interacting with the environment in absence of any supervision is performed by Reinforcement Learning (RL).
Motivated by its recent success~\cite{acer_app2, acer_app1} coupled with its ease of use, we use ACER~\cite{acer} to learn an RL agent giving the final saliency map.
Our approach is equally capable of learning the optimal perturbing sequence on a single image, a class of images or a whole dataset of images.
This makes our approach flexible in terms of scalability and generalization.
To the best of our knowledge, we are the first to propose a RL based strategy for explaining AI made decisions and we term our approach as \textbf{R}einforcement \textbf{Ex}planation \textbf{L}earning (RExL).

We perform experiments on three popular benchmarks in Computer Vision, ImageNet \cite{imagenet_cvpr09}, PASCAL VOC \cite{pascal_voc} and MSCOCO \cite{coco}. For each of these datasets, we use VGG$16$ \cite{vgg} and ResNet$50$ \cite{resnet} as the base models i.e. the classifiers who's decisions are to be explained.  We compare our results with recent black box and white box methods. We show that RExL does not hurt the performance and provides competitive results. In fact it performs better than the baselines in a few cases. Finally we also show the speedup given by RExL compared to other the black box techniques.

\section{Related Work}

There have been a lot of work in Explainable AI \cite{bbm2, guided_backprop, infth1, Petsiuk2018RISE, Fong_2019_ICCV, gradcam} recently. These have used different techniques to explain the decisions taken by Deep Learning models.
Some of them~\cite{infth1, infth2} leverage the use of Information Theory to maximize the mutual information between the compressed representation and the decision. \cite{infth1}, in addition, also reduces the mutual information between the explanation and the input to ensure that the explanation chooses the minimal subset of input features.
These involve complex objective functions which are difficult to use.

Backpropagation Based Methods generate the important measure of a pixel by backpropagating the output of a deep neural network back to the input space using gradients or their variants.
Some of them \cite{bbm1, bbm2} use the derivatives of a class score with respect to the image as an importance score. Guided Backprop \cite{guided_backprop} and DeConvNet \cite{zeiler_fergus} modify the backpropogation scheme for ReLU while Excitation Backprop \cite{excitation_backprop} ensures that the sum of the attribution signal is unity.
Integrated gradients \cite{int-grad} additionally accumulate gradients along a path from base image to input image.
Activation based methods like CAM \cite{cam} (and its variants like Grad-CAM \cite{gradcam} and NormGrad \cite{norm}) use a linear combination of activation values across the convolutional layers. Grad-CAM  and NormGrad uses gradients information to weigh the maps. These methods can generate saliency maps for any activation layer in the network but it is found that later layers, i.e. layers closer to the output generate more informative saliency maps.

These techniques achieve interpretability by making changes to a white-box model or utilising the internals of it. Thus, they cannot always be generalised over all model architectures. This demands for a need of black box techniques that can be used for any base model. LIME \cite{lime} is a black box approach which draws random samples around each instance to be explained and fits an approximate local linear decision model in the vicinity of the input. For complex non-linear classifiers, LIME fails to achieve good performance. Its dependence on superpixels leads to inferior saliency maps. Perturbation based methods perturb the inputs (blur some pixels, or add some noise etc) and measure the response of the model to these. \cite{fine_grained} is such a method that also fine grains gradients to avoid adversarial examples. RISE \cite{Petsiuk2018RISE} is a perturbation based method that generates a large number of random masks and applies them on the image. It then generates the saliency map as the weighted sum of these masks. The weights are the class scores predicted by the model after the mask is applied. It has been successful in generating decent saliency maps but it uses large number of masks (around $5000$) which makes it computationally expensive and time taking. Meaningful perturbations \cite{Fong_2017_ICCV} and Extremal perturbations \cite{Fong_2019_ICCV} solve optimization objective on single images to generate explanations but these involve several iterations of every images and take time.

Reinforcement Learning is increasingly been adopted in standard computer vision tasks like Object Detection \cite{rl_objdet, rl_objdet2, rl_objdet3}, Visual Question Answering \cite{rl_vqa}, Action Recognition \cite{claster} etc. 
TRPO \cite{trpo}, PPO \cite{ppo} and DQN \cite{dqn} are the most commonly used RL algorithms. TRPO and PPO are on-policy while DQN is off-policy. There have been lots of developments on Actor Critic Methods \cite{a3c, acer, sac}. These include both off-policy algoritms \cite{sac} and on-policy methods \cite{a3c}.
There has been some work done \cite{rl_perturbation} where RL agents are used to perturb the inputs to study model behaviour. In most cases, the agent chooses a particular perturbation on the input and then gets rewards from the outputs of a downstream model for the perturbed input.
We follow a perturbation based explanation technique that uses RL to learn an optimal mask and generate accurate saliency maps quickly.

\section{Proposed Approach}
We now describe the proposed method which leverages Reinforcement Learning for getting saliency based explanation of a convnet classifier where at any time, most convincing evidences are collected depending on what evidences have already been collected.

\subsection{Problem Definition}
\label{subsec:prob_def}
Formally, we deal with a dataset $\mathcal{D}$ of images where each image $\mathcal{I}$ can contain one or more objects.
The base model or the model we want to explain provides the probabilities $p_c, \forall c\in \{1, 2, \cdots, C \}$ indicating the presence of at least one object of class $c$ among $C$ classes.
For each object in the image, the explanation generation task is to output importance values for each pixel (or sub-images/patches) of the image.
An explanation is good if gradually removing pixels from the image in order of decreasing importance score confuses the base model.
This will be manifested by a sharp drop of the probability of the object compared to the case when all pixels were present.

We cast explanation generation as a Reinforcement Learning (RL) problem in a Markov Decision Process (MDP) setting.
An MDP has a set of actions $\mathcal{A}$, a set of states $\mathcal{S}$ and a reward function $R:\mathcal{S}\times\mathcal{A} \rightarrow \mathbb{R}$.
At any time instant $t$, the agent selects an action $a_t \in \mathcal{A}$, executes it and as a result, the state changes from $s_t$ to $s_{t+1}$. From the environment, the agent receives 1) a scalar reward $R(s_t, a_t)$ and 2) the next state in the form of an observation.
The action to be taken by the agent is governed by a policy $\pi:\mathcal{S} \rightarrow \mathcal{A}$ that takes in environment states at any point of time and outputs a probability distribution over a set of actions.
The goal of an RL agent is to learn an optimum policy $\pi^*$ such that the sum of rewards through an episode is maximized.
Next we outline the actions, states and rewards of the proposed RL setup.

\noindent{\textit{\textbf{States}}:}
For an image of size $H \times W$, the number of possible masks naturally, is $2^{H \times W}$ as each pixel can be either masked or not.
The state at any time instant $t$ is the masked image where masked pixels are replaced with random noise sampled uniformly in the range of the pixel intensity values. 
For a reasonably sized image, the state-space and thus the search-space of size $2^{H \times W}$ can be huge.
Thus we relax the state-space size by discretizing the image into a $K \times K$ grid, ($K$ being $7$ or $14$).

\noindent{\textit{\textbf{Actions}}:}
Denoting the probability given by the base model for the image to belong to class $c \in \{1, 2, \cdots, C \}$ as $p_c$, the actions taken by the agent aims to reduce $p_c$ to a minimum.
The agent chooses regions to construct the mask sequentially so that the base model provides a low probability or a low score for the masked image to belong to that class.
It should be noted that theoretically, the agent can choose an already masked region also.
However, a well trained agent would find no incentive in performing such an action as the probability value will not reduce much (if at all, it reduces) as a result of masking an already masked region.
For an image divided as $K\!\times\!K$ grid, the number of possible actions is naturally $K^2$.
There seems to be an incentive in terms of training computation to stop the search early by limiting the agent to take only a preset number of actions.
However, this may restrict the exploration ability of the agent.
Thus, to strike a balance, we allow the agent to take as many actions as there are number of grids \textit{i.e.}, $K^2$.

\noindent{\textit{\textbf{Rewards}}:}
The reward function quantifies the worth of the agent's effort for successfully masking important image regions.
A successful masking operation in turn, provides a good saliency map explaining the decision behind predicting the presence of a class in the image.
This is manifested by the drop in probability of a class present in it.
As a natural choice, thus, the reward is taken to be the negative of the probability given by the base model for the class when the perturbed image is passed through the base model.
It is given by,
\begin{equation}
    \label{eq:reward_function}
    R(s_t, a_t) = - p_c^{(t)}
\end{equation}
where $s_t$ is the state and $a_t$ is the action executed at timestep $t$.
$p_c^{(t)}$ is the probability of the image to belong to class $c$, when the masked image at timestep $t$ is passed through the base model.
To maximize the discounted return $G = \mathbb{E} \big[ \sum_t \gamma^t R(s_t, a_t) \big]$, the agent will have to choose the regions such that the classification score of the class drops as quickly as possible thereby minimizing the deletion score.
The discount factor $\gamma \in (0, 1]$ determines the present value of future rewards.
As $\gamma \rightarrow 0$, the agent tries to maximize only the immediate rewards, while a value close to $1$ means the agent becomes farsighted.

\subsection{Solution Strategy}
\label{subsec:sol_strategy}
The goal of the agent is to choose relevant regions of the input image to mask such that the probability or the score of the masked image drops as quickly as possible during the interactions with the environment until termination.
The core of the problem is to find the policy function $\pi$ mapping states to actions that guides the agent's decision making process.
We use a recently proposed Actor Critic with Experience Replay (ACER)~\cite{acer} which is an improvement to the traditional actor-critic algorithm that is more sample efficient.

The policy function of the agent is represented by a multi-layer neural network with learnable parameters $\theta$.
An episode in an RL setting is composed of all the state action pairs from start to the terminal action and is denoted as $\tau = (s_1, a_1, s_2, a_2, \cdots s_T, a_T)$ where $T$ is the number of steps taken before the termination of the episode.
The agent learns the optimal policy by maximizing the expected return $J(\theta)$ over the possible trajectories.
\begin{equation}
	\label{eq:rl_obj}
    J(\theta) = \mathbb{E}_{p_{\theta}(\tau)}\big[\sum_{t} \gamma^{t} R(s_t, a_t) \big] = \mathbb{E}_{p_{\theta}(\tau)}\big[\sum_{t} \gamma^{t}r_{t}\big]
\end{equation}
$R(s_t, a_t)$ is abbreviated as $r_{t+1}$ above to reduce clutter.
The objective function in Equation~(\ref{eq:rl_obj}) can be directly optimized using standard optimization techniques like SGD but suffers from high variance especially under small samples resulting in very fragile convergence.
With the help of the state and action value functions, the Actor Critic algorithms~\cite{konda2000actor} overcome this limitation.
While the state value function $V^{\pi}(s)$ gives the expected return given the agent is at state $s$ and follows the policy $\pi$, the action value function $Q^{\pi}(s,a)$ is the expected return given that the agent performs an action $a$ at state $s$ and follows the policy $\pi$ thereafter.
The modified objective in terms of the advantage $A^{\pi}$ is given by,
\begin{equation}
    J(\theta) = \mathbb{E}_{p_{\theta}(\tau)}\big[\sum_{t} A^{\pi_{\theta}}_{t}\big] \qquad \text{where } A^{\pi_{\theta}}_t = Q^{\pi_{\theta}}(s_t, a_t) - V^{\pi_{\theta}}(s_t)
    \label{eq:rl_obj_adv}
\end{equation}

The basic objective function (Equation~(\ref{eq:rl_obj_adv})) optimized by the Actor-Critic methods is inherently on policy. ACER's objective contains the both off-policy and on-policy components. For the off-policy component, it uses Retrace algorithm \cite{retrace} to estimate the action value function and further uses importance sampling. The on-policy part is very similar to the basic objective shown in Equation (\ref{eq:rl_obj_adv}). 

\begin{figure}[t!]
\centering
\includegraphics[width=140mm]{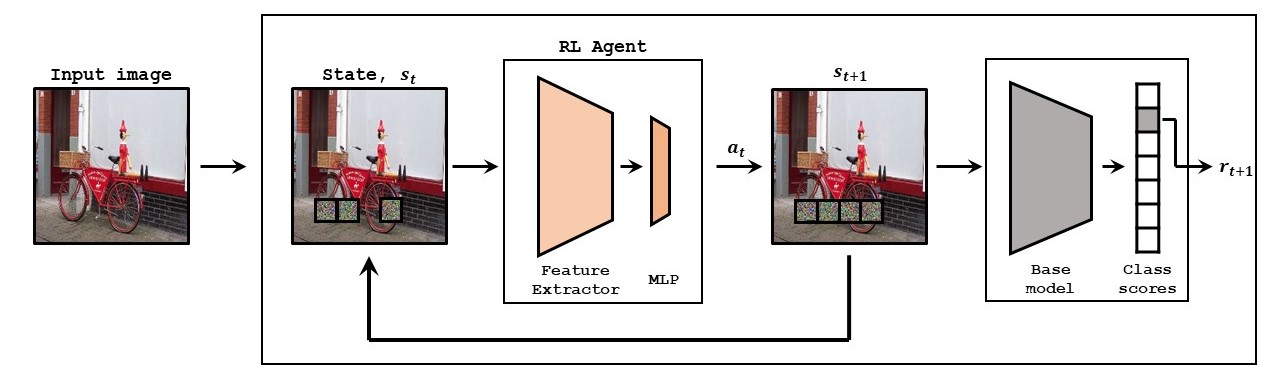}
\caption{\small \textbf{Training the Reinforcement Learning Agent for Explaining Image Classification Model}. The state, $s_t$, at time $t$ is a partially masked image. The RL Agent, which consists of a fixed feature extractor and a learnable MLP, chooses an action $a_{t+1}$. The action is the region of the image that needs to be masked. The result of the action is the next state i.e. $s_{t+1}$. The base model evaluates $s_{t+1}$ to provide the probability of the class being explained which is used to construct the reward $r_{t+1}$.
}
\label{fig:meth}
\end{figure}

An illustration of our proposed approach RExL is provided in Figure~(\ref{fig:meth}).
We provide a saliency map based explanation to the classification decision made by the base model on an image.
The image is given as the input.
The RL agent of RExL consists of a pretrained feature extarctor and a multi-layer perceptron (MLP).
The feature extractor remains fixed during the training and the learnable MLP is the policy network to predict the action.
At every step, the agent observes the perturbed image which constitutes it's state and predicts the next action.
The action is a specific region in the image to mask.
Once the image is perturbed or the action is executed, the base model gives the classification score for the class to be explained.
This confidence score is used as reward during training (ref. Equation~(\ref{eq:reward_function})) and also to generate the explanations as saliency maps.
The perturbed image becomes the next state and the loop continues until termination.

RExL gives the flexibility to train a single agent to explain all images of a dataset (\emph{dataset specific}), multiple agents each one expertly trained to explain a single class (\emph{class specific}) or extremely fine-grained agents that are experts in explaining a single image (\emph{image specific}).
We train the dataset specific agents by using the whole training partition of the dataset in which the base model was trained.
A class specific agent is trained on a single class of training images. 
Finally, an image specific agent is trained only on the image to be explained.
The explainability of image specific agents are very good but this is not scalable.
Although dataset specific agent is scalable and performs good for images containing mostly single objects, the performance may degrade for a complex dataset with images having multiple instances from possibly multiple classes.
As our experiments show, the best trade-off between performance and scalability can be seen for class specific agents.

\subsection{Inference}
\label{subsec:inference}

After a model is trained, it is used to generate saliency maps.
Given an image, the agent sequentially deletes a block of the grid from it and incrementally builds the saliency map.
Let block $b_t$ of an image is perturbed as a result of the action $a_t$ taken by the agent at time $t$ and the change in classification confidence from the previous state is $\delta_t$.
The saliency weight assigned to block $b_t$ is $\sum\limits_{i=t}^{T} \lambda^{i-t}\delta_t$, where $\lambda \in [0,1]$ is the cumulating factor that gives the due credit to previous deletions that may have resulted in a greater drop in probability at a later stage.
A $\lambda$ value equal to $0$ is the trivial case where only the grid that is deleted most recent gets all the credit.
Finally, the map is normalized so that the saliency weights sum to $1$.
\section{Experiments}
\label{sec:exp}
\noindent{\textit{\textbf{Datasets and Evaluation Metrics}}:}
\label{subsec:dataset}
We evaluate the performance of our proposed method on three benchmark datasets namely ImageNet \cite{imagenet_cvpr09}, PASCAL VOC $2007$ \cite{pascal_voc} and MSCOCO $2014$ \cite{coco}. Following the standard practice, we have reported the results from the \emph{val}, \emph{test} and \emph{val} splits respectively.
For all the three datasets, we use pretrained VGG$16$ \cite{vgg} and ResNet$50$ \cite{resnet} as base models. For ImageNet, we used the trained models provided by Pytorch Models Zoo \footnote[1]{https://pytorch.org/vision/stable/models.html}.
For PASCAL VOC and MSCOCO, we used the same multilabel models as RISE and made public by~\cite{excitation_backprop}.

We follow~\cite{Petsiuk2018RISE} and use the causal deletion and insertion metrics to compare the performance with different approaches~\cite{Petsiuk2018RISE, Fong_2019_ICCV, norm, gradcam}.While GradCAM~\cite{gradcam} and NormGrad~\cite{norm} are white box techniques that make use of the intermediate activations and gradients of the network, RISE~\cite{Petsiuk2018RISE} and Extremal Perturbations~\cite{Fong_2019_ICCV} are black box techniques that only study the input-output relationship of the model. RISE is a sampling based method and samples thousands of random masks and weighs them based on their performance. Extremal Perturbations on the other hand is a learning based method which solves an optimization objective for each image to obtain the corresponding mask.

Deletion metric removes pixels from the image gradually according to the importance weight given by the saliency map and measures the classification score.
The deletion score is the area under the curve (AUC) of these classification scores with percentage of pixels removed.
Similarly the insertion score evaluates the performance inversely \textit{i.e.}, uncovering highly blurred image regions gradually according to the importance weights and getting the AUC.
While a low deletion score is indicative of better explanation, in case of insertion score the opposite is true.
Though, RExL follows the deletion philosophy while training, the insertion score for it is also competitive.
We posit that an agent trained with an insertion strategy (\textit{i.e.}, starting with a blank canvas and learning to choose relevant regions to uncover) would be equally useful and we keep it as a possible future work.
While our primary aim was to improve on the inference time of black box model, we first show that RExL does not hurt the performance, infact in some cases, it performs better than even white box methods. Thereafter we provide a runtime analysis showing the significant speedup given by us compared to similar black box XAI strategies.

\noindent{\textit{\textbf{Implementation Details}}:}
\label{subsec:implement}
The policy feature extractor can be any good pretrained feature extractor and is independent from the base model. We use ResNet$50$ pretrained on the dataset the RL agent being trained on as the feature extractor for the policy.
The learnable MLP consists of two hidden layers with $256$ and $128$ units respectively.
We fix the value of $K$ as $7$ and the length of the episode, thus, will be $7^2= 49$.
We trained ACER \footnote{Implementation: https://stable-baselines.readthedocs.io/en/master/modules/acer.html} for $2 \times 10^6$ steps when training class specific and $10^7$ steps when training dataset specific and used discount factor, $\gamma = 1$ throughout our experiments. Also for all the inferences, we have used $\lambda=1$ unless explicitly specified.
As dataset specific setting is a single agent to explain all the classes in the dataset, information about the class to be explained was passed as an one-hot encoding along with the image during training.
We used NVIDIA GeForce RTX $2080$Ti GPUs for training all our models.
Next we discuss the experimental results for all the three variants of RExL \textit{i.e.}, dataset specific (\emph{RExL-DS}), class specific (\emph{RExL-CS}) and image specific (\emph{RExL-IS}).
Due to limitations in space, we provide the comparison over deletion metric only.
Additional results including insertions scores, more implementation details and source codes are provided in the appendix.

\subsection{Results and Discussions}

\begin{figure}[t!]
\centering
\includegraphics[width=140mm]{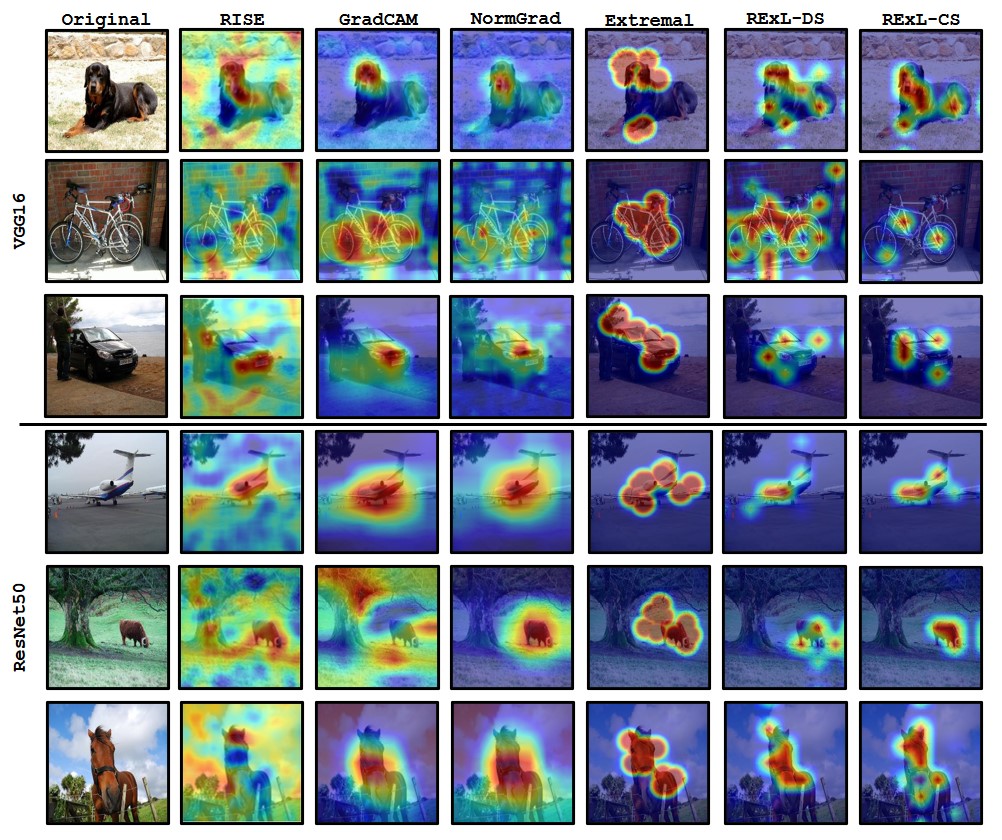}
\caption{\small \textbf{Comparison of the saliency maps}: The top three rows show saliency maps for $3$ random samples from VOC$2007$ when RExL agent is applied on VGG$16$ base model. Likewise the bottom $3$ rows correspond to ResNet$50$ base model. The leftmost column shows the original image. Rest of the column headings show the approach used to generate the respective saliency maps. The results show the superiority of the proposed approach over the black box and white box approaches as well as the specificity of the maps by the RExL-CS approach compared to the RExL-DS approach.}

\vspace{-10pt}
\label{fig:all}
\end{figure}

\begin{table*}[t!]
    \centering
    \begin{tabular}{|c||c|c||c|c||c|c|}
        \hline
        \multirow{2}{*}{Method} & \multicolumn{2}{c ||}{ImageNet} & \multicolumn{2}{c ||}{PASCAL VOC 2007} & \multicolumn{2}{c |}{COCO 2014} \\ \cline{2-7}
         & VGG16 & ResNet50 & VGG16 & ResNet50 & VGG16 & ResNet50\\
         \hline
         GradCAM \cite{gradcam} & 0.109 & 0.123 & \textbf{0.275} & 0.310 & \textbf{0.225} & 0.261 \\
         \hline
         NormGrad \cite{norm} & \textbf{0.086} & 0.126 & 0.285 & 0.349 & 0.232 & 0.303 \\
         \hline 
         RISE~\cite{Petsiuk2018RISE} & 0.098 & \textbf{0.108} & \textbf{0.275} & 0.290 & 0.240 & 0.227 \\
         \hline 
         Extremal \cite{Fong_2019_ICCV} & 0.125 & 0.126 & 0.355 & 0.357 & 0.292 & 0.299 \\
         \hline 
         RExL-DS & 0.130 & 0.122 & 0.342 & \textbf{0.243} & 0.284 & \textbf{0.200} \\
         \hline
    \end{tabular}
    \caption{\small \textbf{Comparison of Deletion scores on Imagenet, PASCAL VOC 2007 and COCO 2014 Datasets}. RExL-DS outperforms white box methods on VOC $2007$ and COCO $2014$ for ResNet$50$ base model. For a less strong VGG$16$ base model, RExL-DS is competitive.}
    \label{tab:entire_dat_del}
\end{table*}

\begin{wraptable}{r}{0.4\linewidth}
    \centering
    \begin{tabular}{|c||c|c|}
        \hline
         Method & VGG16 & ResNet50 \\
         \hline
         RExL-DS & 0.342 & 0.243 \\
         \hline
         RExL-CS & 0.292 & 0.216 \\
         \hline 
    \end{tabular}
    \caption{\small \textbf{Comparison of the Deletion scores of RExL-DS and RExL-CS on VOC2007 test split}. The deletion score is improved for the more specific RExL-CS agent compared to RExL-DS for both the base models.}
    \label{tab:class_spec}
\end{wraptable}

Table~(\ref{tab:entire_dat_del}) shows a comparative evaluation of the deletion scores on the three benchmark datasets with two different base models explaining the classes present in the images. RExL-DS performs better when explaining a stronger model like ResNet$50$.
For PASCAL VOC $2007$ and COCO $2014$ datasets, RExL-DS outperforms all the baselines (including the white box approaches like GradCAM and NormGrad) while for ImageNet, its performance is at par (slightly better than most of the other baselines).
With VGG$16$, the performance is at par with the competing methods.
The comparatively higher deletion scores of RExL-DS for VGG$16$ is intuitive as this being an inferior base model the reward for a right choice of image region to delete may not always be right.

As we go to class specific RExL agents (ref Table~(\ref{tab:class_spec})), our approach becomes more competitive for VGG$16$ base model while further improvement is seen in case of ResNet$50$ making RExL-CS the new SOTA for PASCAL VOC.
In case of Imagenet, due to the presence of a large number of classes, we focused on explaining the top $10$ classes for which VGG16 and ResNet50 classifiers give best performance in terms of top-$1$ accuracy.
The average deletion score of the RExL-CS agents trained on these $10$ classes for ResNet50 base model is as low as $0.111$ while RISE being the closest with a value of $0.149$.
Likewise, for VGG16, the proposed approach gives a value of $0.139$ with RISE being close with a deletion score of $0.149$.
The improvement over RExL-DS shows the flexibility of our approach as well as the expertise imbibed by the class specific models compared to one single agent trained to explain all the classes in the dataset.
Additional results on both Imagenet and PASCAL VOC are provided in the appendix.

Figure~(\ref{fig:all}) shows a comparison between the saliency maps generated by different baseline methods and the two variants of RExL.
Targeted saliency and much less noise of the RExL maps make them qualitatively better compared to the others.
While extremal perturbation~\cite{Fong_2019_ICCV} generates less noisy maps, its dependence on manually set area of relevant regions can harm if the object of interest does not conform to this preset value (pronounced in the top row for VGG$16$ and penultimate row for ResNet$50$).
This is especially detrimental for RISE where the probability of the masked regions in the image is fixed likewise.
RExL, on the other hand, learns to mask regions that is important for making the base model decide the class without constraining the area to be prespecified.
Improvement by our method is also observed when the specificity of the agent increases and it becomes more expert \textit{i.e.}, when we go from dataset-specific to class-specific agents.
This is especially prominent for the image of the cycle in case of VGG$16$ and for the image of the cow in case of ResNet$50$.

Although RExL-IS is the extremely specific, time to train image specific models for all images is also high.
As a result, we don't specifically evaluate RExL-IS on all images and provide quantitative results.
Instead, we do qualitative comparison with existing approaches on a few images along with a comparison of the deletion scores.
Due to space limitation we provide additional qualitative examples including RExL-IS comparisons in the appendix.

\subsection{Comparison of Inference Time}
\label{subsec:runningtime}

The major advantage of and motivation for RExL is the targeted fast search of relevant image regions without degrading the quality of the explanations.
In this section we compare the inference times of RExL with that of the state-of-the-art black box explanation techniques \textit{e.g.}, RISE and the extremal perturbation approaches.We do not compare the running time of RExL with white box techniques like GradCAM and NormGrad because they are much quicker but come at the cost of peeking into the model. Our aim is to bring about considerable speedup in the inference time of a black box explanation technique without hurting the performance.

For comparing the running times, we use a computer with one NVIDIA GeForce GTX $1080$Ti, ryzen $2700$x CPU and $16$GB DDR$4$ $3000$MHz RAM.
Table (\ref{tab:running_time}) shows the average running times over $100$ images.
RExL provides almost $10\times$ speedup over both the approaches for VGG$16$ base model.
The improvement in speed is almost $5\times$ and $10\times$ respectively over RISE and extremal perturbation in case of ResNet$50$ base model.
RExL-CS is comparatively fast than RExL-DS as RExL-DS gets an additional class encoding as input.
The faster running time with comparable or better performance shows the merit of our RL based explainable AI approach.

\begin{table*}[t!]
    \centering
    \begin{tabular}{|c||c|c||c|c||c|c|}
        \hline
        \multirow{2}{*}{Method} & \multicolumn{2}{c ||}{ImageNet} & \multicolumn{2}{c ||}{PASCAL VOC} & \multicolumn{2}{c |}{MSCOCO} \\ \cline{2-7}
         & VGG16 & ResNet50 & VGG16 & ResNet50 & VGG16 & ResNet50\\
         \hline
         RExL-CS & 1.461 & 1.494 & 1.748 & 1.488 & 1.738 & 1.528 \\
         \hline
         RExL-DS & 1.869 & 1.872 & 1.754 & 1.532 & 1.800 & 1.519 \\
         \hline
         RISE \cite{Petsiuk2018RISE} & 14.618 & 8.853 & 20.408 & 8.370 & 20.397 & 8.392 \\
         \hline
         Extremal \cite{Fong_2019_ICCV} & 13.722 & 15.975 & 19.114 & 16.298 & 19.118 & 16.298 \\
         \hline
    \end{tabular}
    \caption{\small \textbf{Comparison of Running Times (in seconds) while generating saliency maps}. RExL-DS and RExL-CS are compared with the related black box methods (RISE and Extremal Perturbations). Both variants of RExL outperforms the competing methods by a large margin.
    }
    \label{tab:running_time}
\end{table*}

\begin{figure}
    \begin{minipage}[c]{0.55\linewidth}
    \centering
        \includegraphics[width=\textwidth]{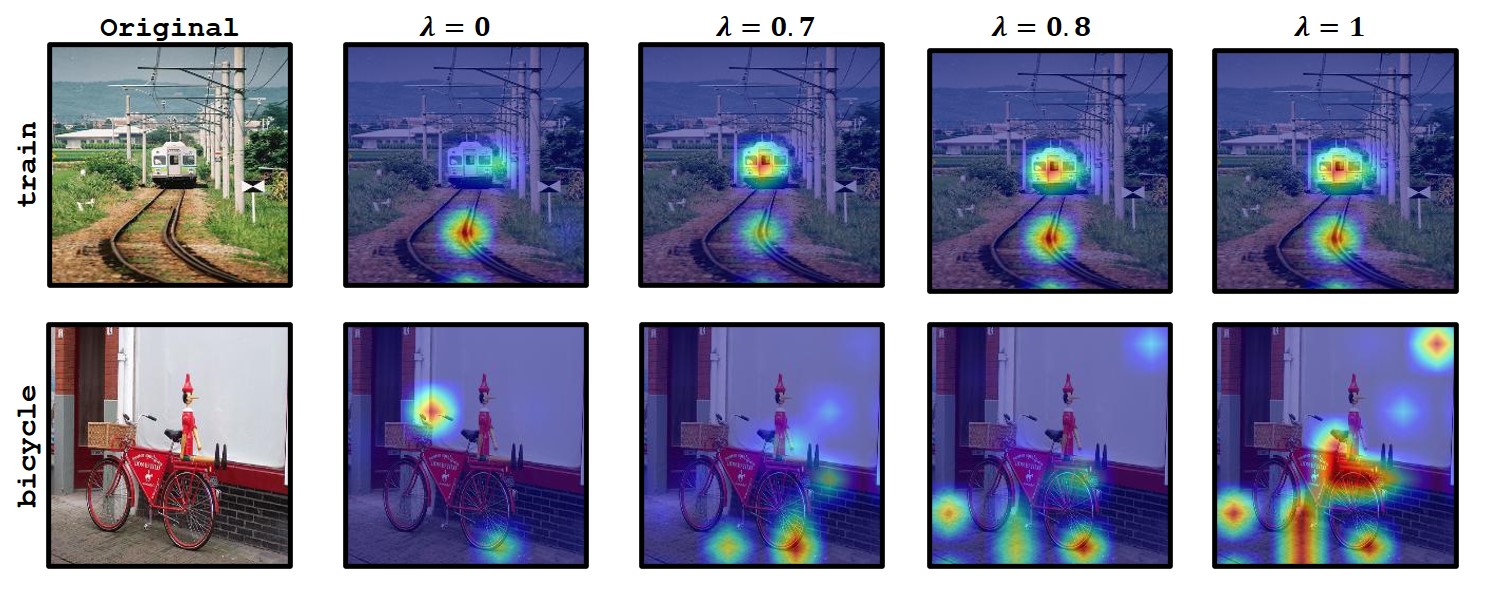}
        \caption{\small \textbf{Comparison between the saliency maps produced by RExL-DS for different values of the cumulating factor $\boldsymbol\lambda$}. The $\lambda$ values are put on the top. The leftmost column shows the original image from the PASCL VOC $2007$.}
        \label{fig:lambda}
    \end{minipage}
    \hspace{0.1cm}
    \begin{minipage}[c]{0.35\linewidth}
    \centering
        \begin{tabular}{|c||c|c|}
            \hline
             Inference & VGG16 & ResNet50 \\
             \hline
             $\lambda = 0$ & 0.362 & 0.290 \\
             \hline
             $\lambda = 0.7$ & 0.349 & 0.266 \\
             \hline 
             $\lambda = 0.8$ & 0.347 & 0.260 \\
             \hline 
             $\lambda = 1$ & 0.342 & 0.243 \\
             \hline 
        \end{tabular}
    \captionof{table}{\small \textbf{Comparison of the deletion scores for the different values of the cumulating factor} $\boldsymbol\lambda$ during inference on VOC$2007$ test split using RExL-DS.}
    \label{tab:inf_tech}
    \end{minipage}
    
\end{figure}

\subsection{Ablation on Cumulating Factor}

To see the effect of the cumulating factor $\lambda$ while assigning credit to the chosen image regions we experimented with different values of it for RExL-DS on the PASCAL VOC dataset.
Table~(\ref{tab:inf_tech}) shows the comparison.
Though the values are close, one important observation is that giving the entire credit at a certain instant to the region that is deleted most recently (\textit{i.e.}, $\lambda\!=\!0$) hurts the performance.
As $\lambda\!=\!1$ provides the best performance, we went ahead in experimenting with this value for all the datasets.
Figure~(\ref{fig:lambda}) shows sample saliency maps with different $\lambda$ values applied on the same image.
For a large object, naturally, a large portion of the image is responsible for the base model to get its decision and this is rightly captured when credit assignment is delayed instead of immediate.
\section{Conclusions}
\label{sec:conclusion}

We have presented a reinforcement learning based model for explaining deep classifiers.
In contrast to methods that approximate an exhaustive space, we propose to formulate XAI as an intelligent search and direct an agent to efficiently choose and accumulate evidences responsible behind AI made decisions.
Our model is dataset, category and image specific striking a perfect balance of model specificity and training burden.
Ours is a plug and play black box approach that is applicable to any classifier without requiring to modify the internals of them.
We report encouraging results in three benchmark datasets achieving $5$ to $10$ times speed-up over related black box XAI approaches.

\bibliographystyle{plain}
\bibliography{ref}

\newpage
\section*{Appendix}
\appendix
\section{Implementation Details}

In this section, we discuss all the details that are needed to reproduce our results as well as the scores obtained for the baseline methods. These parameters can be categorized as follows:

\subsection{Environment and Agent Parameters}

The mask is discretized into a $K\times K$ grid as discussed in section 3 of the main paper. For our experiments we choose $K = 7$. Naturally, the number of steps in an episode as discussed will be $49$. For VOC $2007$ and COCO $2014$, we used VGG$16$ and ResNet$50$ trained by \cite{excitation_backprop}. These models are trained on BGR images loaded with pixel values in $[0, 255]$ and then mean corrected to $[104.01, 116.67, 122.68]$ respectively. For ImageNet, we use the pretrained models provided by Pytorch model zoo~\footnote{\url{https://pytorch.org/vision/stable/models.html}} and used the standard normalization practices described there. The random noise that is used to mask the image is sampled from the same distribution as that of the image. For example in case of VOC or COCO, the noise has a mean $[104.01, 116.67, 122.68]$.

The RL agent consists of two parts, a feature extractor which is fixed during training and a learnable MLP. The feature extractor used is a pretrained ResNet$50$. So, for VOC$2007$, we will be using the ResNet$50$ module that was pretrained on VOC$2007$ by \cite{excitation_backprop}. The learnable MLP consists of two hidden layers of $256$ and $128$ units each. The activation functions at these hidden units is ReLU. The output layer predicts the catergorical distribution over actions and hence contains $49 (K^2)$ output neurons. 

\begin{wraptable}[10]{l}{0.4\linewidth}
\centering
\begin{tabular}{|p{3cm}|c|}
    \hline
    Parameter & Value \\
    \hline
    \hline
    discount factor, $\gamma$ & 1.0 \\
    \hline
    Number of environment steps per update & 490 \\
    \hline
    weight for loss of Q value & 1.0 \\
    \hline
    RMS prop alpha & 0.9 \\
    \hline
    learning rate & 0.0001 \\
    \hline
\end{tabular}
\caption{RL training hyperameters}
\label{tab:rl_param}
\end{wraptable}

During inference, as reported in section 4.4, $\lambda = 1$ gives the best performance so we report the results using $\lambda = 1$. RExL agents are trained on the respective partitions of the train split and following standard practices \cite{Petsiuk2018RISE, Fong_2017_ICCV, Fong_2019_ICCV, excitation_backprop} tested on the val, test and val split of ImageNet, VOC and COCO.

\subsection{RL training parameters}

We used ACER\footnote{\url{https://stable-baselines.readthedocs.io/en/master/_modules/stable_baselines/acer/acer_simple.html\#ACER}} to train the policy. The parameters that we used during training are in Table (\ref{tab:rl_param}). We used the default hyperparameters as mentioned in the official documentation\footnote{\url{https://stable-baselines.readthedocs.io/en/master/modules/acer.html}} for the rest of the parameters.

\subsection{Baselines}

For RISE we used the official implementation\footnote{\url{https://github.com/eclique/RISE}} with the hyperparameters discussed in \cite{Petsiuk2018RISE}. For GradCAM and NormGrad, we used the publically available implementation\footnote{\url{https://github.com/ruthcfong/TorchRay/tree/normgrad}}. Finally for Extremal Perturbations, we used the implementation in the official repository\footnote{\url{https://github.com/facebookresearch/TorchRay}} using contrastive rewards and area parameter of $0.12$ while generating explanations.
\begin{table*}[t!]
    \centering
    \begin{tabular}{|c||c|c||c|c||c|c|}
        \hline
        \multirow{2}{*}{Method} & \multicolumn{2}{c ||}{ImageNet} & \multicolumn{2}{c ||}{PASCAL VOC 2007} & \multicolumn{2}{c |}{COCO 2014} \\ \cline{2-7}
         & VGG16 & ResNet50 & VGG16 & ResNet50 & VGG16 & ResNet50\\
         \hline
         GradCAM \cite{gradcam} & 0.615 & 0.677 & 0.850 & 0.850 & 0.713 & 0.740 \\
         \hline
         NormGrad \cite{norm} & 0.517 & 0.455 & 0.795 & 0.807 & 0.640 & 0.690 \\
         \hline 
         RISE~\cite{Petsiuk2018RISE} & 0.666 & 0.727 & 0.764 & 0.782 & 0.631 & 0.701 \\
         \hline 
         Extremal \cite{Fong_2019_ICCV} & 0.641 & 0.690 & 0.871 & 0.870 & 0.718 & 0.750 \\
         \hline 
         RExL-DS & 0.464 & 0.529 & 0.755 & 0.755 & 0.592 & 0.678 \\
         \hline
    \end{tabular}
    \caption{\small \textbf{Comparison of Insertion scores on Imagenet, PASCAL VOC 2007 and COCO 2014 Datasets}. }
    \label{tab:entire_dat_ins}
\end{table*}

\section{Insertion Scores}

 RExL gets competitive scores for all the datasets even though the preocess of insertion is complementary to our training process. A method that optimizes deletion scores will not necessary also give good insertion scores. For deletion score to be low, the agent will learn the mask that will remove some features that increases the confusion in the model for that class. But inserting the same features may not necessarily lead to a high probability for that class. 
 \begin{wraptable}[11]{r}{0.4\linewidth}
    \centering
    \begin{tabular}{|c||c|c|}
        \hline
         Method & VGG16 & ResNet50 \\
         \hline
         RExL-DS & 0.755 & 0.755 \\
         \hline
         RExL-CS & 0.778 & 0.777 \\
         \hline 
    \end{tabular}
    \caption{\small \textbf{Comparison of the Insertions scores of RExL-DS and RExL-CS on VOC2007 test split}. The deletion score is improved for the more specific RExL-CS agent compared to RExL-DS for both the base models.}
    \label{tab:class_spec_ins}
\end{wraptable}
For example, in ImageNet, there are a lot of classes pertaining to different kinds of dogs. There will be some features (like the some portion of the face) on removal of which, the base model will get confused among all the other dogs. But if we insert this feature on a blank image, the base model will not necessarily classify it correctly. 
 
Naturally this will be more pronounced where the classes are a lot similar to each other. As a result, (ref. Table (\ref{tab:entire_dat_ins})) RExL does not do as well in terms of insertion scores (compared to) on ImageNet but does very well on VOC where the classes are a lot different.
Similar to the deletion score, insertion score for RExL-CS is better than RExL-DS, as seen from Table (\ref{tab:class_spec_ins}). RExL-CS results will be further analyzed in the next sections.

\begin{wraptable}[12]{l}{0.37\linewidth}
    \centering
        \begin{tabular}{|c||c|c|}
            \hline
             Inference & VGG16 & ResNet50 \\
             \hline
             $\lambda = 0$ & 0.783 & 0.785 \\
             \hline
             $\lambda = 0.7$ & 0.771 & 0.769 \\
             \hline 
             $\lambda = 0.8$ & 0.768 & 0.765 \\
             \hline 
             $\lambda = 1$ & 0.755 & 0.755 \\
             \hline 
        \end{tabular}
    \captionof{table}{\small \textbf{Comparison of the insertion scores for the different values of the cumulating factor} $\boldsymbol\lambda$ during inference on VOC$2007$ test split using RExL-DS.}
    \label{tab:inf_tech_ins}
    \end{wraptable}

One interesting observation in Table (\ref{tab:inf_tech_ins}) is about the general trend that shows insertion scores are better for lower values of $\lambda$.
This is counterintuitive to the trend seen in case of deletion scores (Table 4 in the main paper).
For an object, the lower values of $\lambda$ credit the later actions more while the larger $\lambda$ value the earlier actions.
The results show that the later actions are more important for insertion scores. These are the actions that are probably removing the significant portions of the object while the earlier actions aim to remove the features that separate similar classes.
We posit that using an \emph{insertion-style} training can lead to agents which will do better in terms of insertion scores too. This is a possible future work.

\newpage

\section{RExL-CS on VOC2007}

For VOC$2007$, we train RExL-CS on all the 20 classes. Saliency maps generated by RExL-CS is compared with the baseline methods for every class. For quantitative comparisons, we use both the deletion scores and insertion scores.  The maps generated by RExL-CS and baselines for every class are also compared qualitatively.

\subsection{VGG16}
\begin{figure}[H]
    \centering
    \includegraphics[width=140mm]{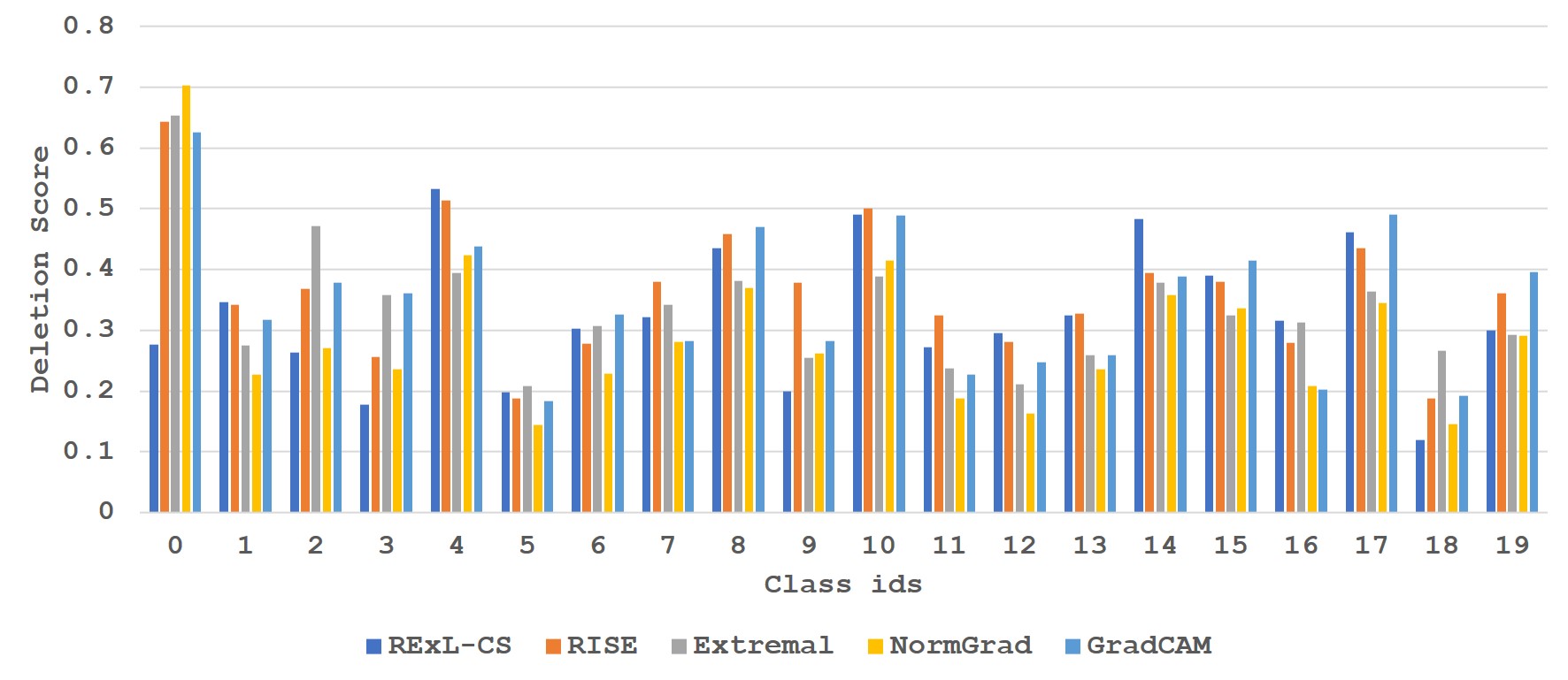}
    \caption{\textbf{Comparison of deletion scores (the lower the better) for VGG$16$ on all classes of VOC$2007$:} RExL-CS gets competitive deletion scores for all the classes. It outperforms all the baselines for 5 classes.}
    \label{fig:del_vgg_pas}
\end{figure}

\begin{figure}[H]
    \centering
    \includegraphics[width=140mm]{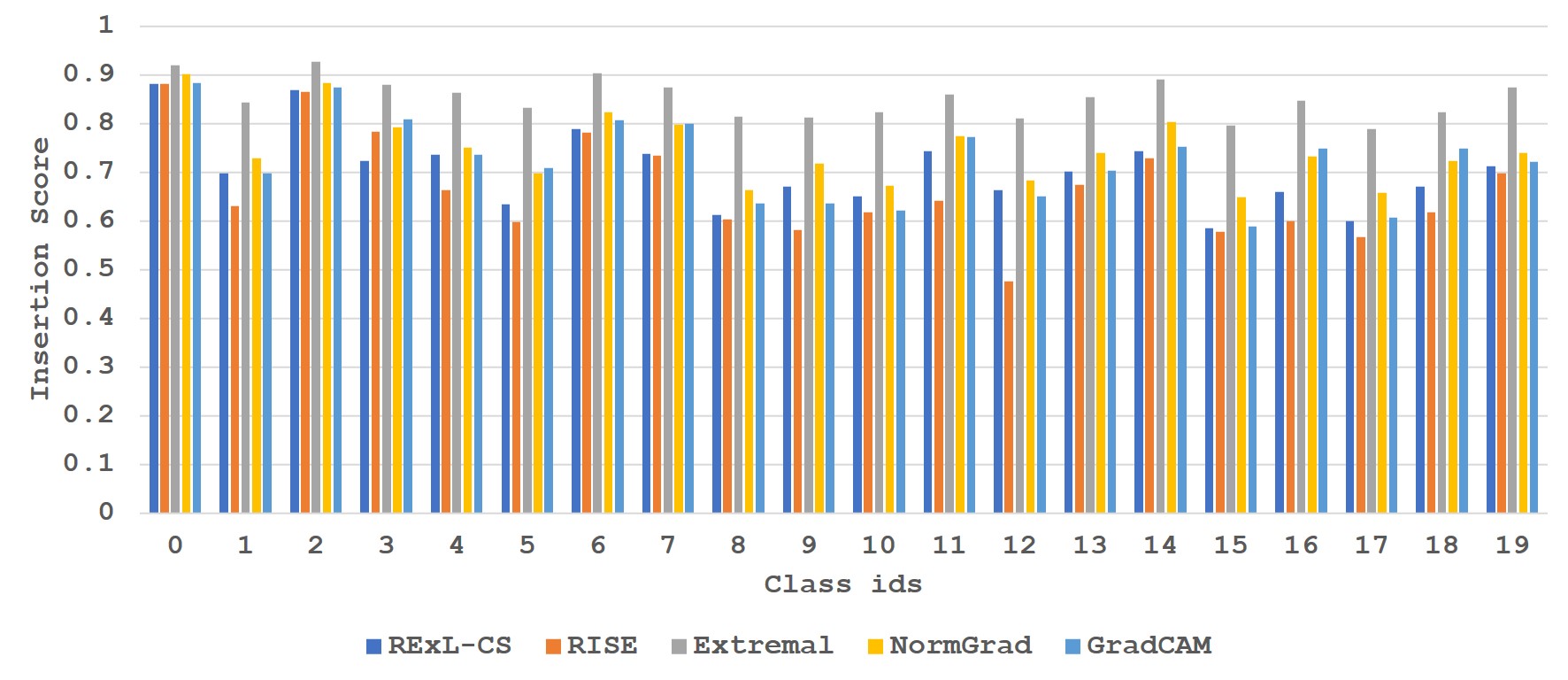}
    \caption{\textbf{Comparison of insertion scores (the higher the better) for VGG$16$ on all classes of VOC$2007$:} Insertion scores of RExL-CS are at par with the baselines. On average it has the better insertion score than RISE.}
    \label{fig:ins_vgg_pas}
\end{figure}

\begin{figure}[H]
    \centering
    \includegraphics[width=130mm]{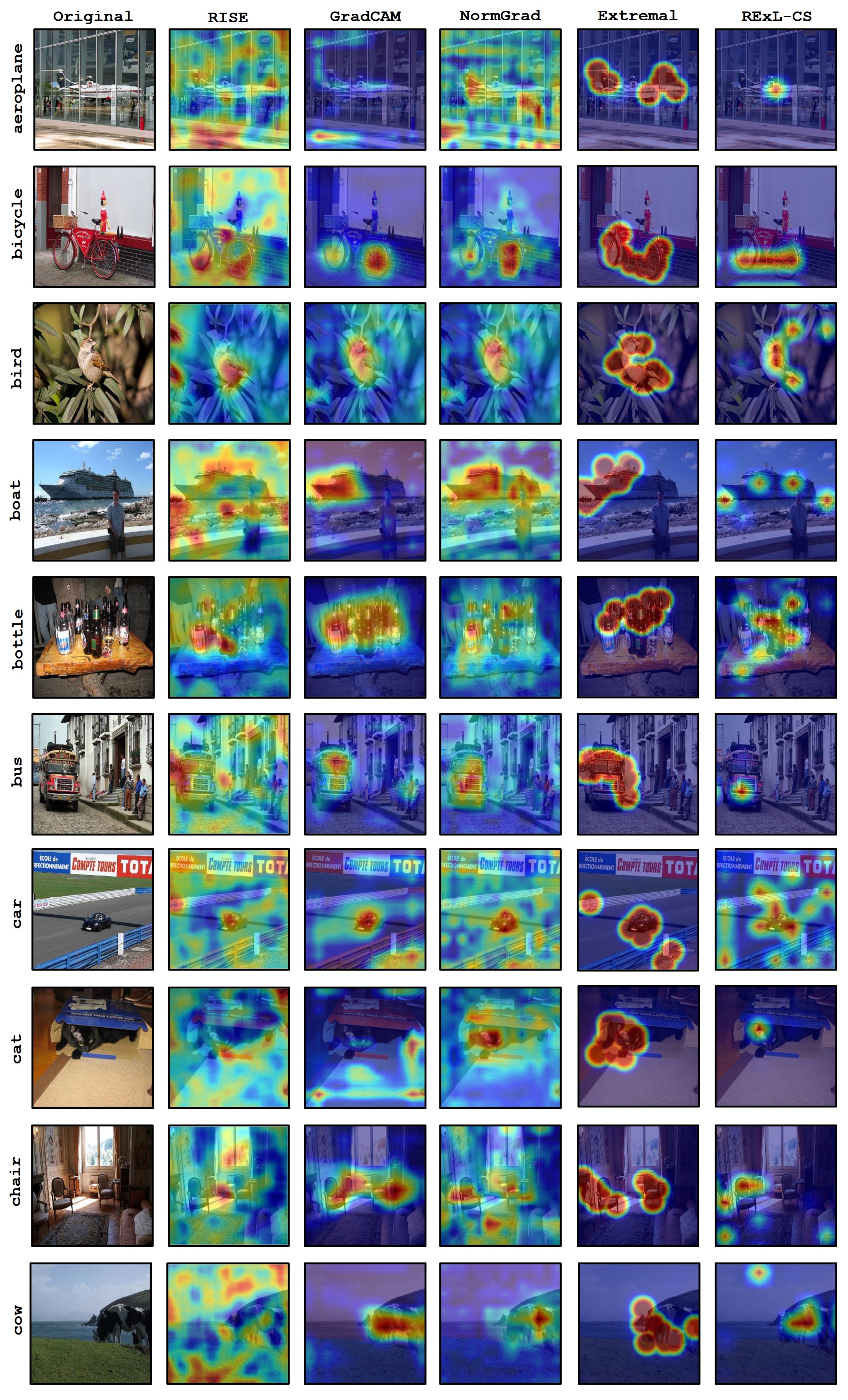}
    \caption{\small \textbf{Qualitative comparison of saliency maps for VGG$16$ on classes $0-9$ of VOC$2007$:} RExL-CS always generates better maps than RISE. Extremal perturbations gives almost equal importance to the entire object but its is difficult to identify the sailient features of the object. RExL-CS is at par with the whitebox methods in most cases and definitely better in some cases like "aeroplane" and "cat".}
    \label{fig:vgg_pas1}
\end{figure}

\begin{figure}[H]
    \centering
    \includegraphics[width=130mm]{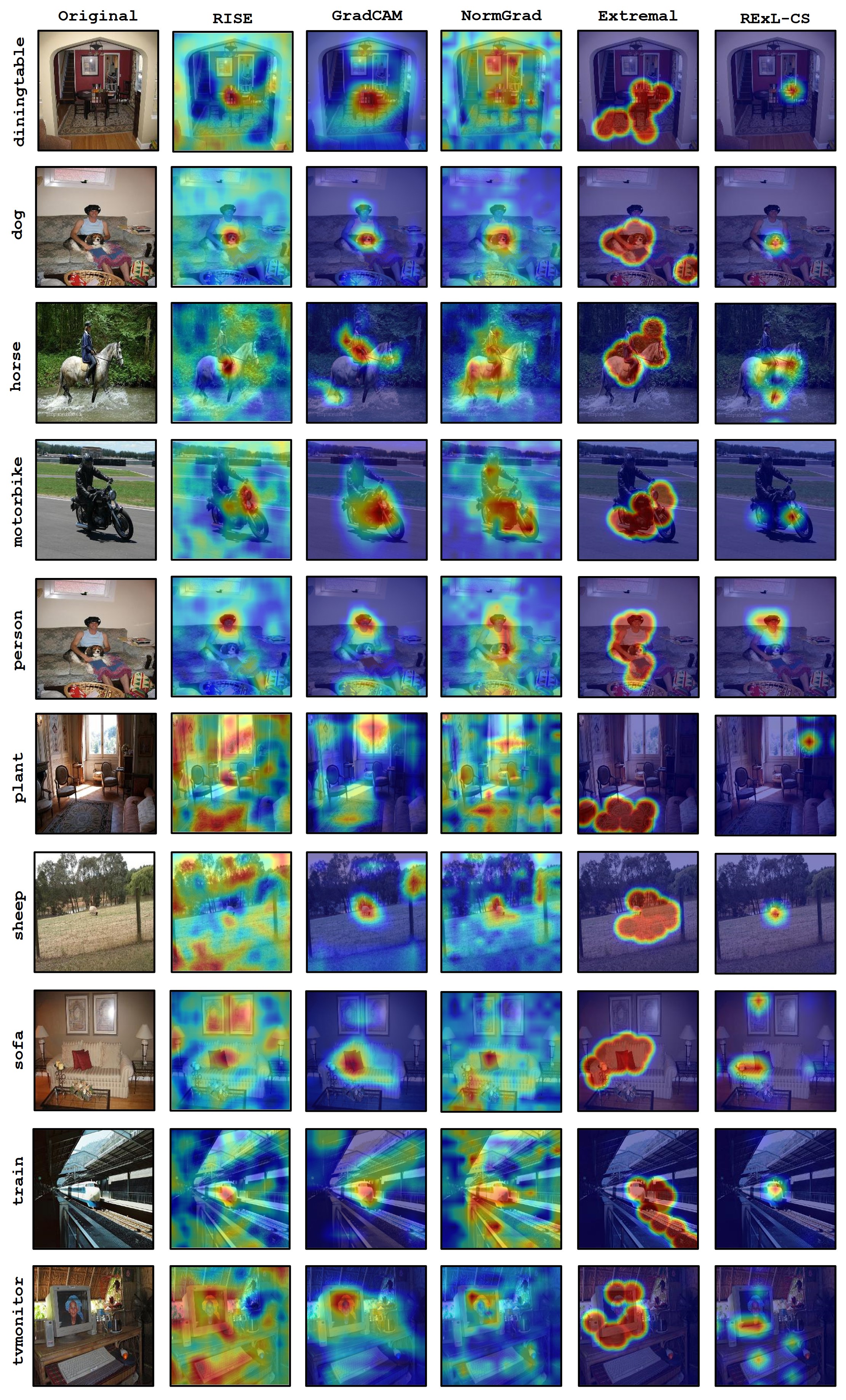}
    \caption{\small \textbf{Qualitative comparison of saliency maps for VGG$16$ on classes $10-19$ of VOC$2007$:} Similar to Figure (\ref{fig:vgg_pas1}), RExL-CS generated better maps than both RISE and Extremal Perturbations. RExL-CS outperforms all the baselines for classes like "plant" and "tvmonitor".}
    \label{fig:vgg_pas2}
\end{figure}

Figure (\ref{fig:del_vgg_pas}) and (\ref{fig:ins_vgg_pas}) show the comparison of class wise deletion and insertion scores among the different methods. Moreover, Figure (\ref{fig:vgg_pas1}) and Figure (\ref{fig:vgg_pas2}) qualitatively compare the saliency maps generated by RExL-CS with the baselines for every class. VGG$16$ gives competitive results for both deletion and insertion scores. The saliency maps produced are also more accurate and concise.

\subsection{ResNet50}
\begin{figure}[H]
    \centering
    \includegraphics[width=140mm]{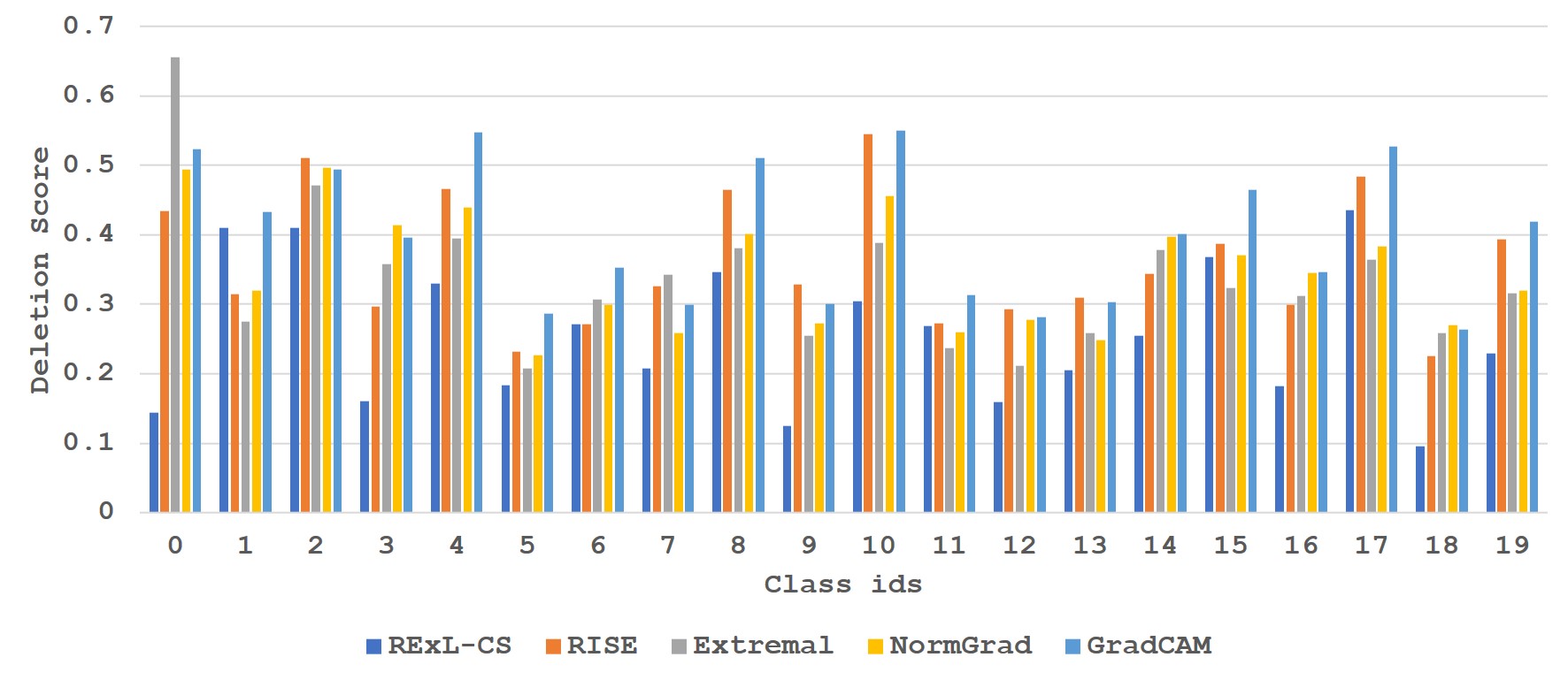}
    \caption{\textbf{Comparison of deletion scores (the lower the better) for ResNet$50$ on all classes of VOC$2007$:} RExL-CS outperforms the baselines most of the times (in $15$ classes). It performs exceptionally well on some classes like "aeroplane" (class id $0$) and this will be further highlighted in the saliency maps in Figure (\ref{fig:res_pas1}).}
    \label{fig:del_res_pas}
\end{figure}

\begin{figure}[H]
    \centering
    \includegraphics[width=140mm]{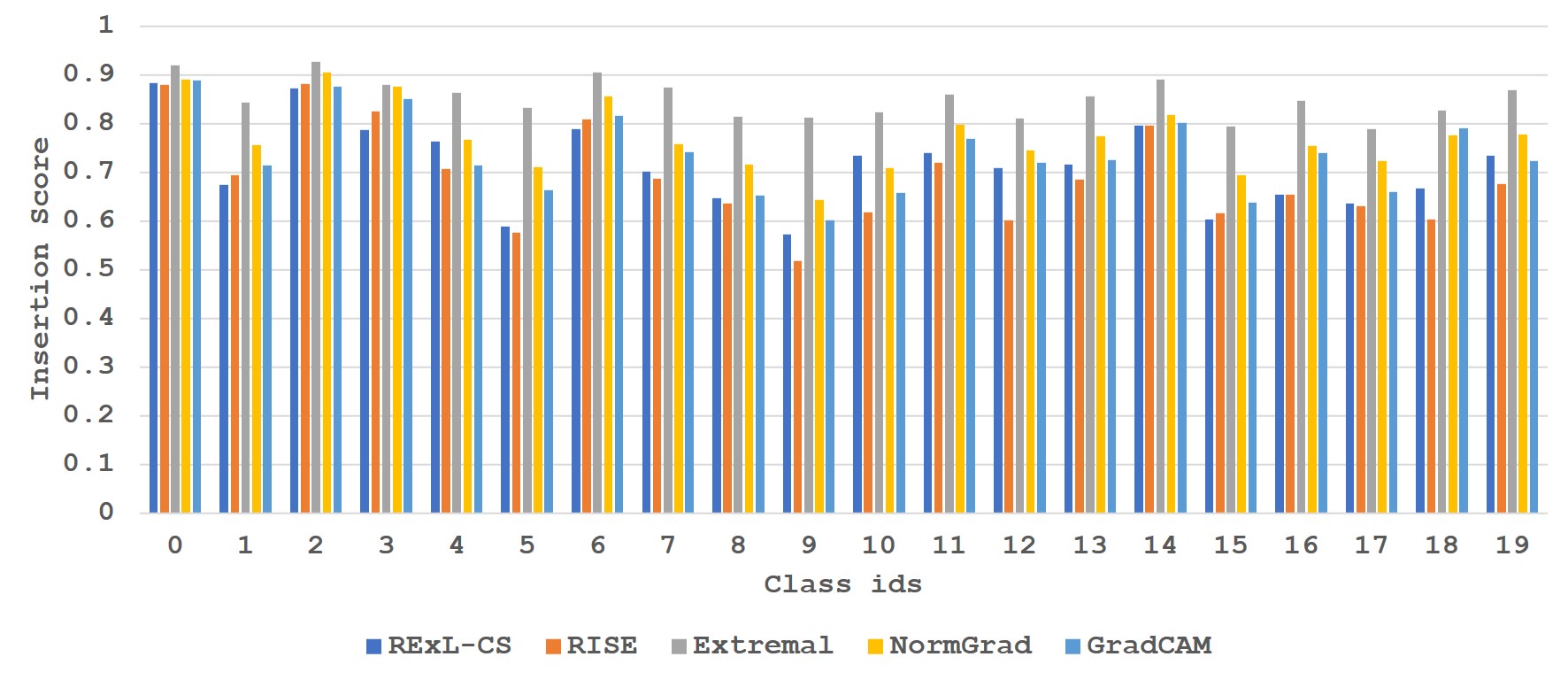}
    \caption{\textbf{Comparison of insertion scores (the higher the better) for ResNet$50$ on all classes of VOC$2007$:} RExL-CS is at par with all the baselines and on average better than RISE.}
    \label{fig:ins_res_pas}
\end{figure}

\begin{figure}[H]
    \centering
    \includegraphics[width=130mm]{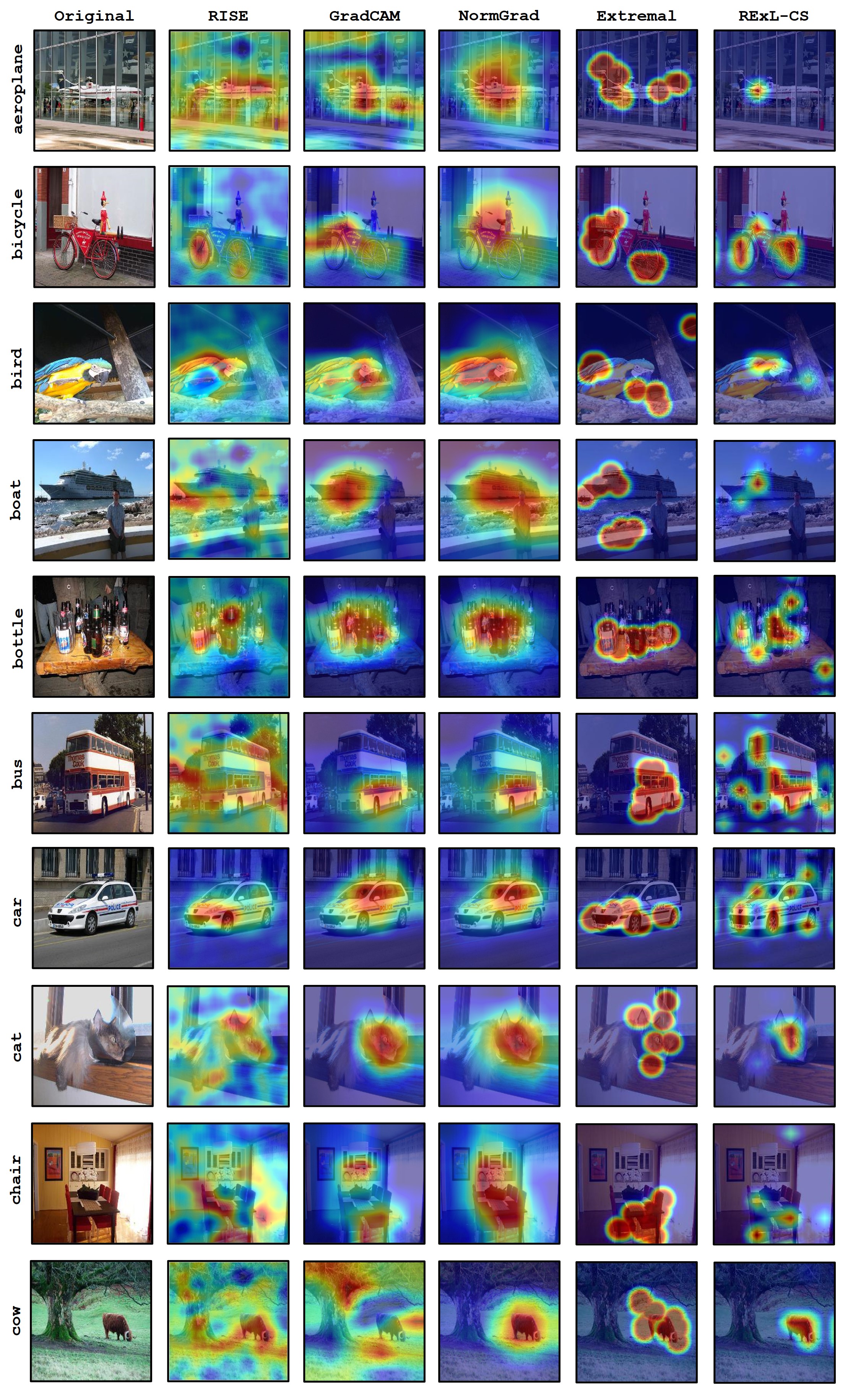}
    \caption{\small \textbf{Qualitative comparison of saliency maps for ResNet$50$ on classes $0-9$ of VOC$2007$:} As expected, the saliency map for "aeroplane" is better than the baseline methods. Moreover, for classes "bicycle", "bus", "chair" and "cow", the maps are much better than the four baselines. For the same input for class "boat", RExL performs way better while using ResNet$50$ than when it was using VGG$16$. }
    \label{fig:res_pas1}
\end{figure}

\begin{figure}[H]
    \centering
    \includegraphics[width=130mm]{ 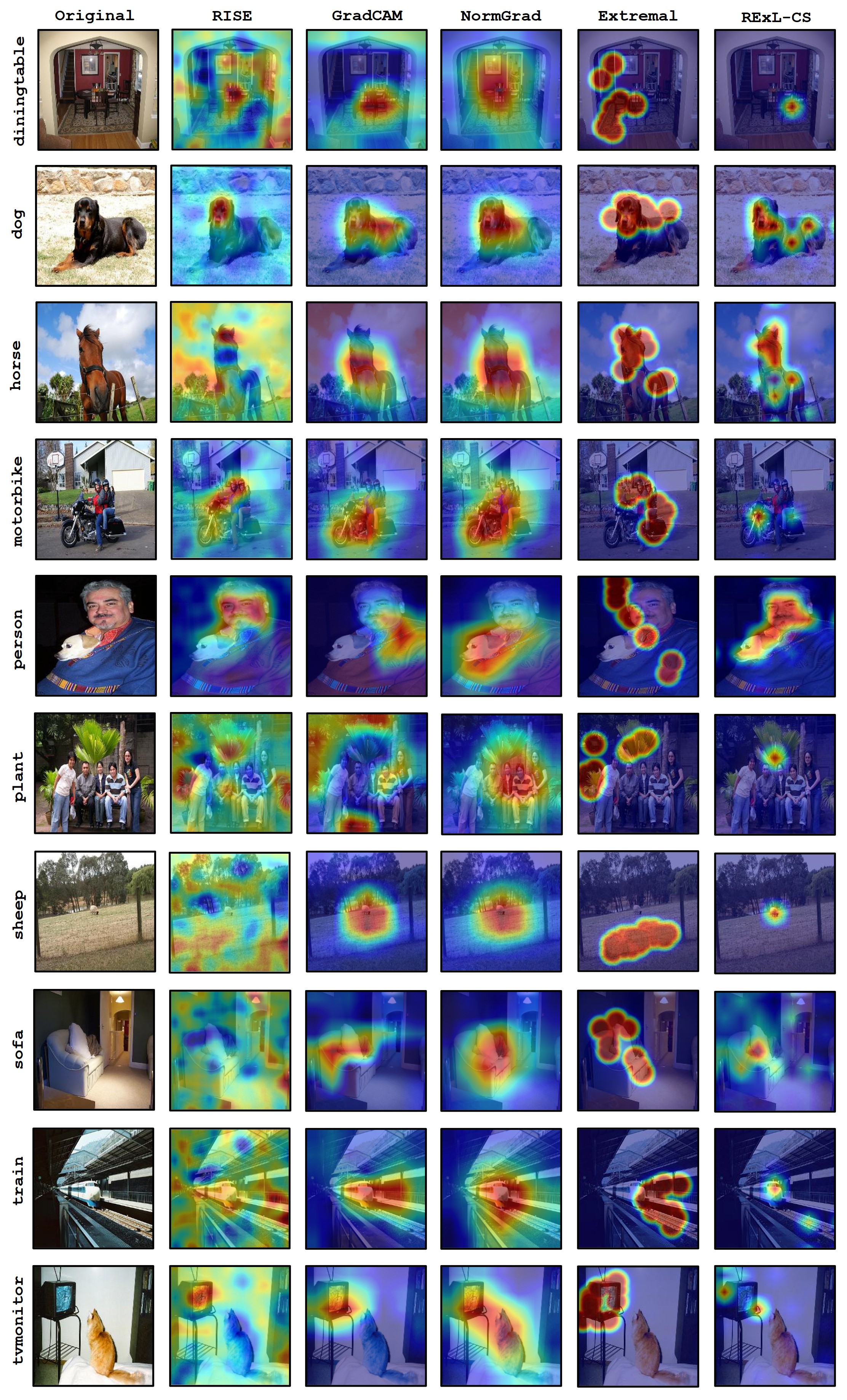}
    \caption{\small \textbf{Qualitative comparison of saliency maps for ResNet$50$ on classes $10-19$ of VOC$2007$:}Like the previous Figure, RExL-CS outperforms the baseline methods in most of the classes. It almost always generates better explanations than the two black box methods and is always at par (and sometimes better) with the white box methods.}
    \label{fig:res_pas2}
\end{figure}

Similarly, RExL-CS agents are trained for all the $20$ classes using ResNet$50$ as the base model. Figure (\ref{fig:del_res_pas}) and Figure (\ref{fig:ins_res_pas}) respectively show the comparison of deletion and insertion scores for every class. As seen in all the results, RExL-CS improves considerable as we use a stronger base model like ResNet$50$. Finally, Figure (\ref{fig:res_pas1}) and Figure (\ref{fig:res_pas2}) contains the saliency maps generated by the different methods for qualitative comparison.

\section{RExL-CS on ImageNet}
We also perform classwise comparison of the saliency maps generated by RExL-CS with the baseline methods on ImageNet. But instead of all the 1000 classes, we show results only on the top 10 classes (as per the classifcation accuracy). Similar to VOC$2007$, we use deletion and insertion scores to compare quantitatively while qualitatively we show the saliency maps produced for a random example from each of the 10 classes. Again we perform the experiments for both the base models.

\subsection{VGG16}
\begin{figure}[H]
    \centering
    \includegraphics[width=140mm]{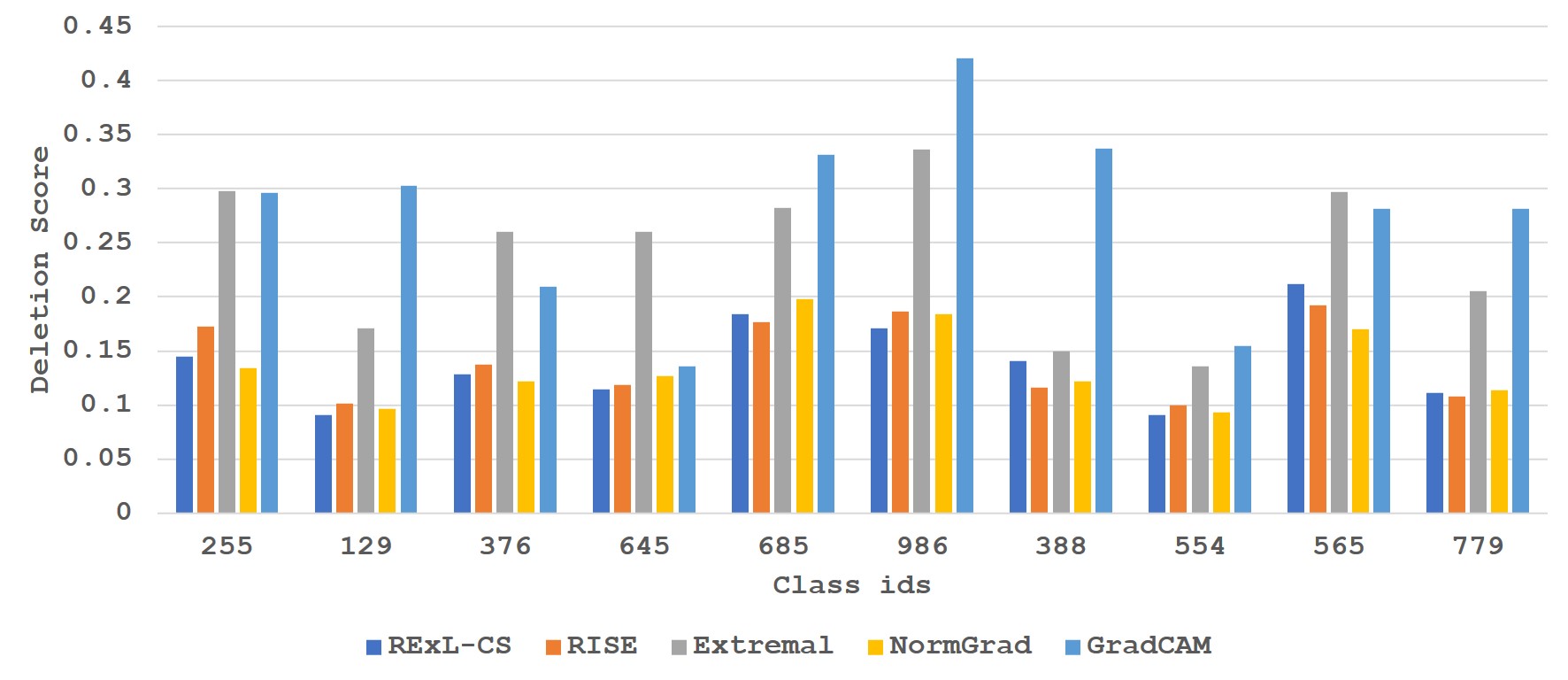}
    \caption{\textbf{Comparison of deletion scores (the lower the better) for VGG$16$ on top 10 classes of ImageNet:} RExL-CS gets competitive deletion scores for all the classes and outperforms in 4 classes.}
    \label{fig:del_vgg_img}
\end{figure}

\begin{figure}[H]
    \centering
    \includegraphics[width=140mm]{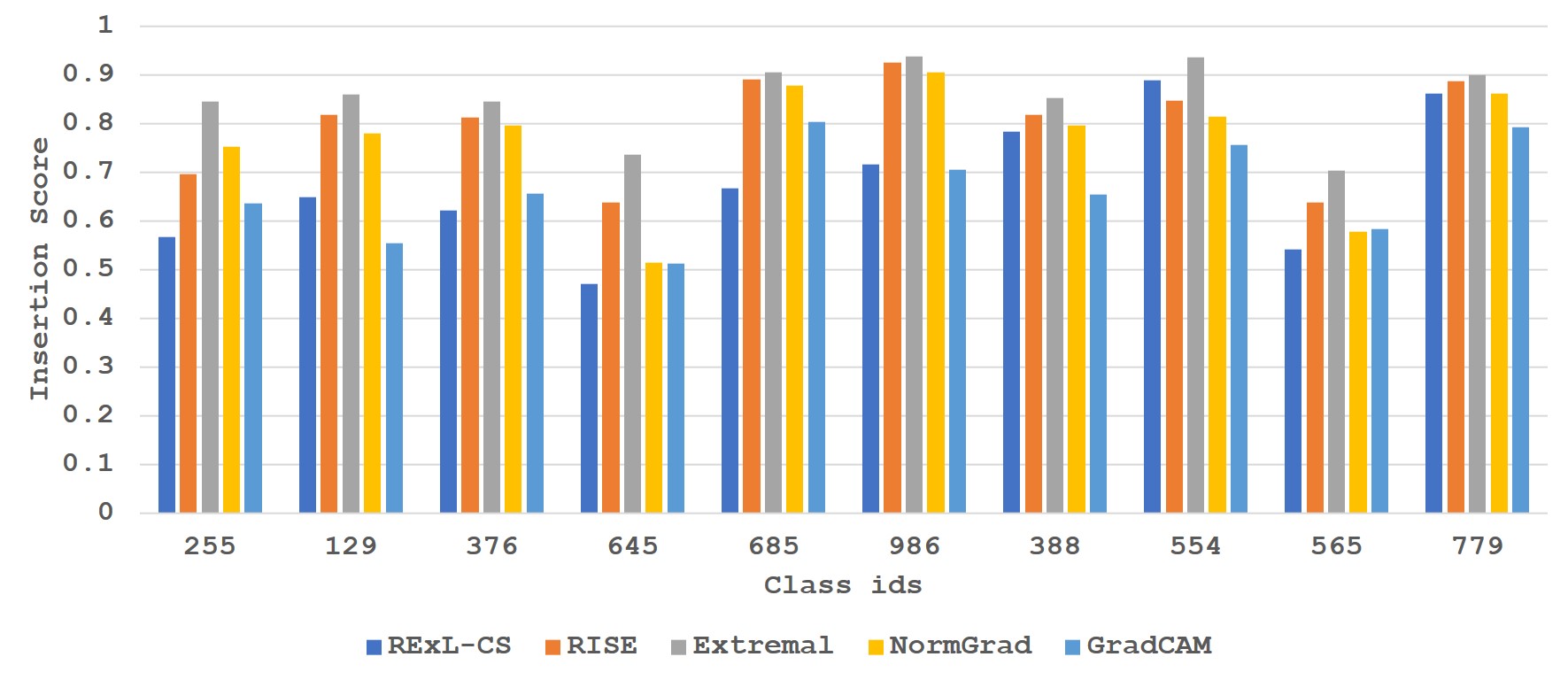}
    \caption{\textbf{Comparison of insertion scores (the higher the better) for VGG$16$ on top 10 classes of ImageNet:} RExL-CS gets competitive insertion scores for all the classes. }
    \label{fig:ins_vgg_img}
\end{figure}

\begin{figure}[H]
    \centering
    \includegraphics[width=130mm]{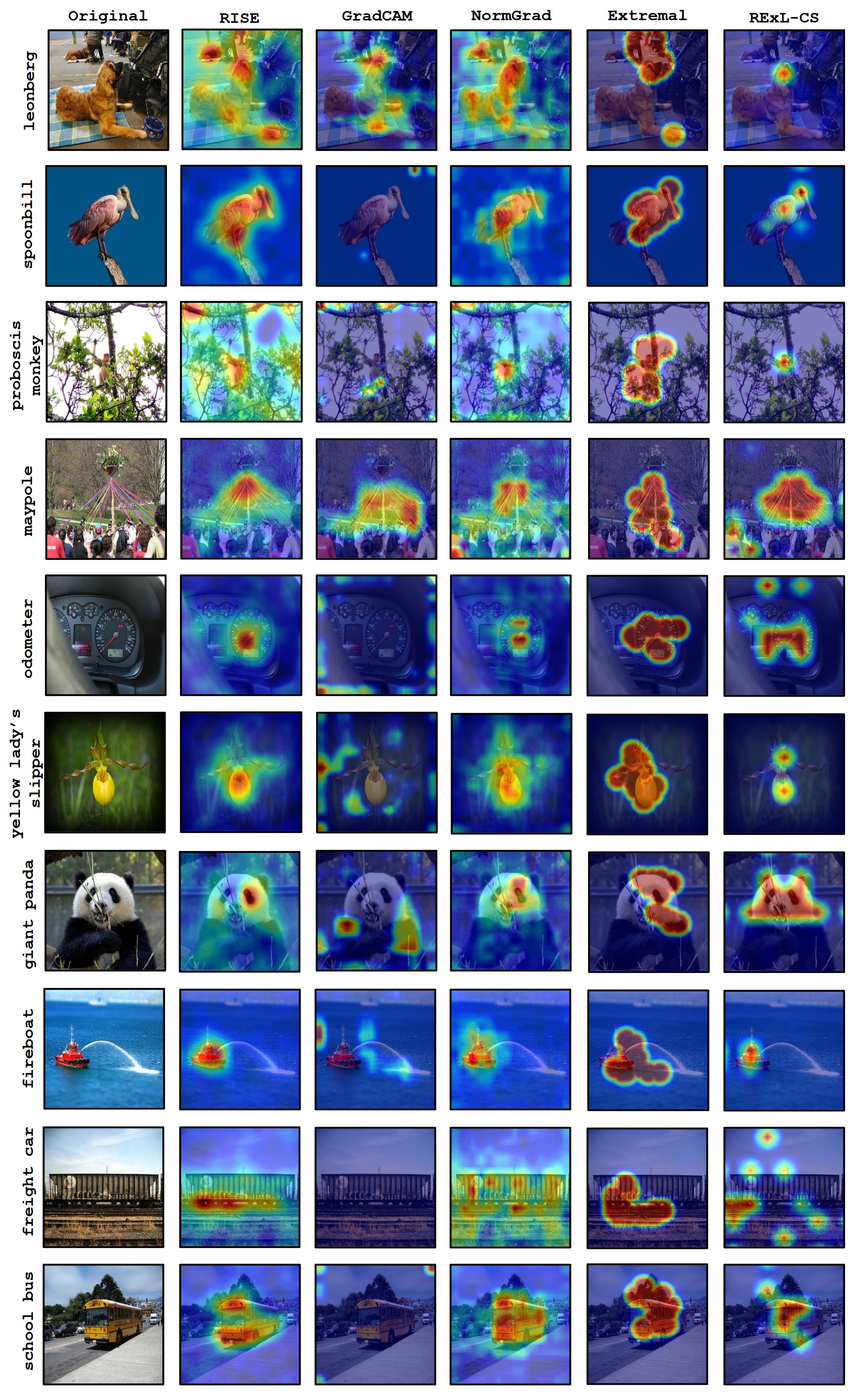}
    \caption{\small \textbf{Qualitative comparison of saliency maps for VGG$16$ on top 10 classes of ImageNet:} RExL-CS always generated better maps than Extremal. In most cases, RExL-CS outperforms RISE as well. GradCAM does not perform well on some classes which can also be reflected from its poor scores on these classes. In several classes like "giant panda",  "leonberg" and "proboscis monkey", RExL-CS generates better maps than all the baselines including the white box methods.}
    \label{fig:vgg_img}
\end{figure}

Figure (\ref{fig:del_vgg_img}) and (\ref{fig:ins_vgg_img}) show the comparison of class wise deletion and insertion scores among the different methods respectively. And, Figure (\ref{fig:vgg_img}) qualitatively compare the saliency maps generated by RExL-CS with the baselines for every class. For VGG16, the scores are at par with the baselines, outperforming the baselines few times.

\subsection{ResNet50}

\begin{figure}[H]
    \centering
    \includegraphics[width=140mm]{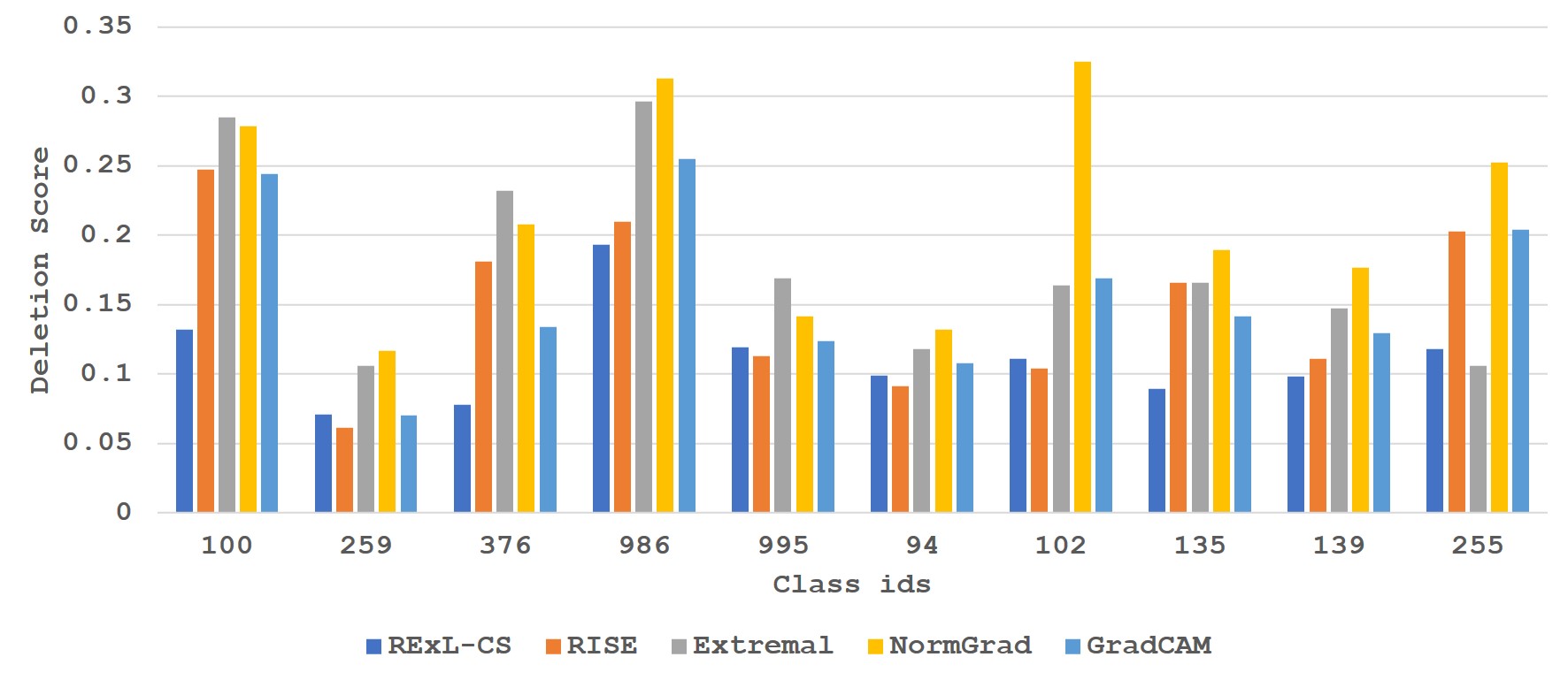}
    \caption{\textbf{Comparison of deletion scores (the lower the better) for ResNet$50$ on top 10 classes of ImageNet:} RExL-CS outperforms the baselines on $5$ classes. Wherever it fails to be the best, it just falls behind by a very small margin and is almost always the second best.}
    \label{fig:del_res_img}
\end{figure}

\begin{figure}[H]
    \centering
    \includegraphics[width=140mm]{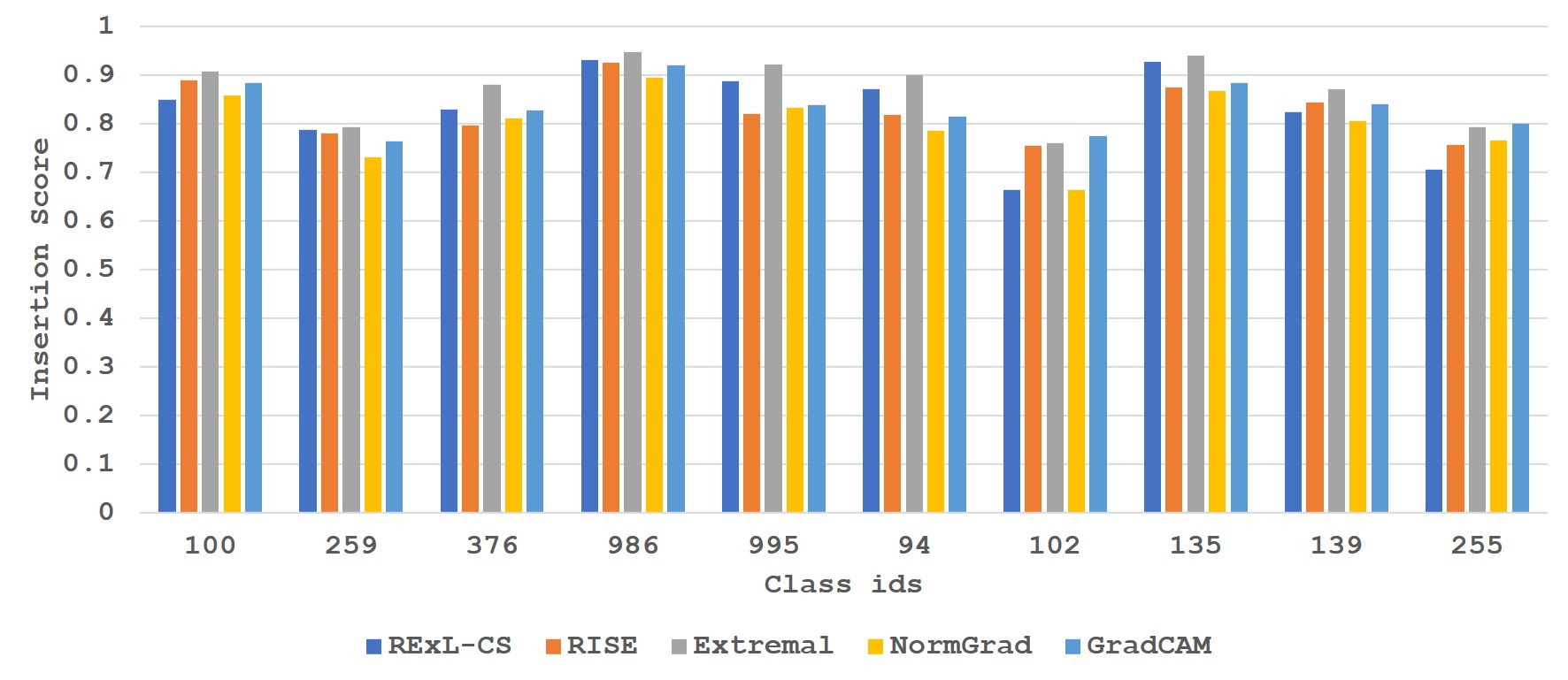}
    \caption{\textbf{Comparison of insertion scores (the higher the better) for ResNet$50$ on top 10 classes of ImageNet:} RExL-CS gets competitive insertion scores for all the classes. On average, it is better than three of the baselines.}
    \label{fig:ins_res_img}
\end{figure}

\begin{figure}[H]
    \centering
    \includegraphics[width=130mm]{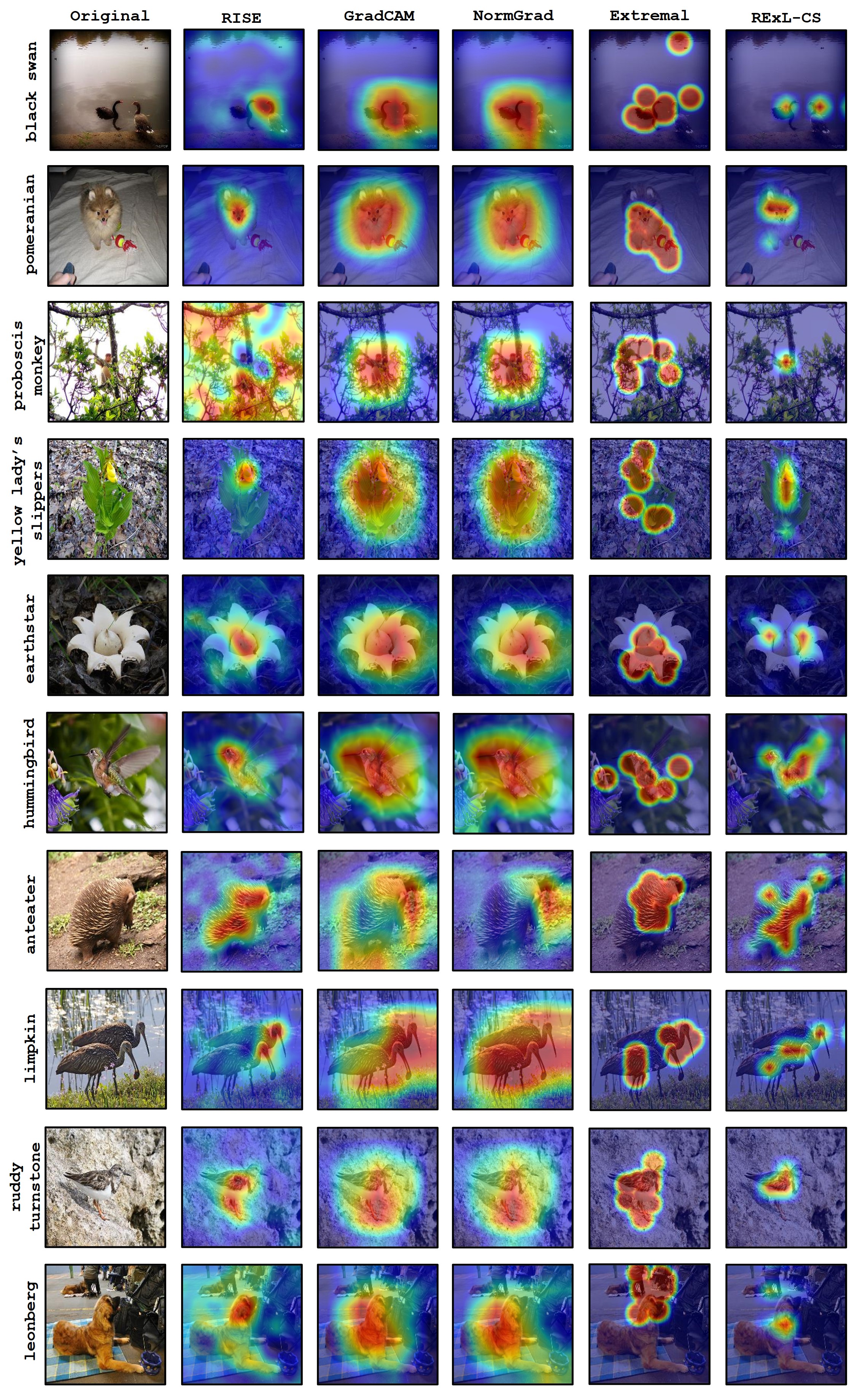}
    \caption{\small \textbf{Qualitative comparison of saliency maps for ResNet$50$ on top 10 classes of ImageNet:} RExL-CS again is always better than Extremal and at par (if not better) than RISE. It is also better than the two white box methods which give a large blob of importance rather than providing any intricate information. For classes "black swan", "proboscis monkey", "hummingbird", "leonberg" and "ruddy turnstone", RExL by far outperforms the baselines.}
    \label{fig:res_img}
\end{figure}

We perform similar comparisons using ResNet$50$ as the base model. Figure (\ref{fig:del_res_img}) and (\ref{fig:ins_res_img}) compare the class wise deletion and insertion scores among the different methods respectively. And, Figure (\ref{fig:res_img}) qualitatively compares the saliency maps generated by RExL-CS with the baselines for every class. As expected, RExL performs better using ResNet$50$ as the base model and this can be illustrated from both the scores and the saliency maps. 

\section{RExL-IS}

\begin{figure}[H]
    \centering
    \includegraphics[width=130mm]{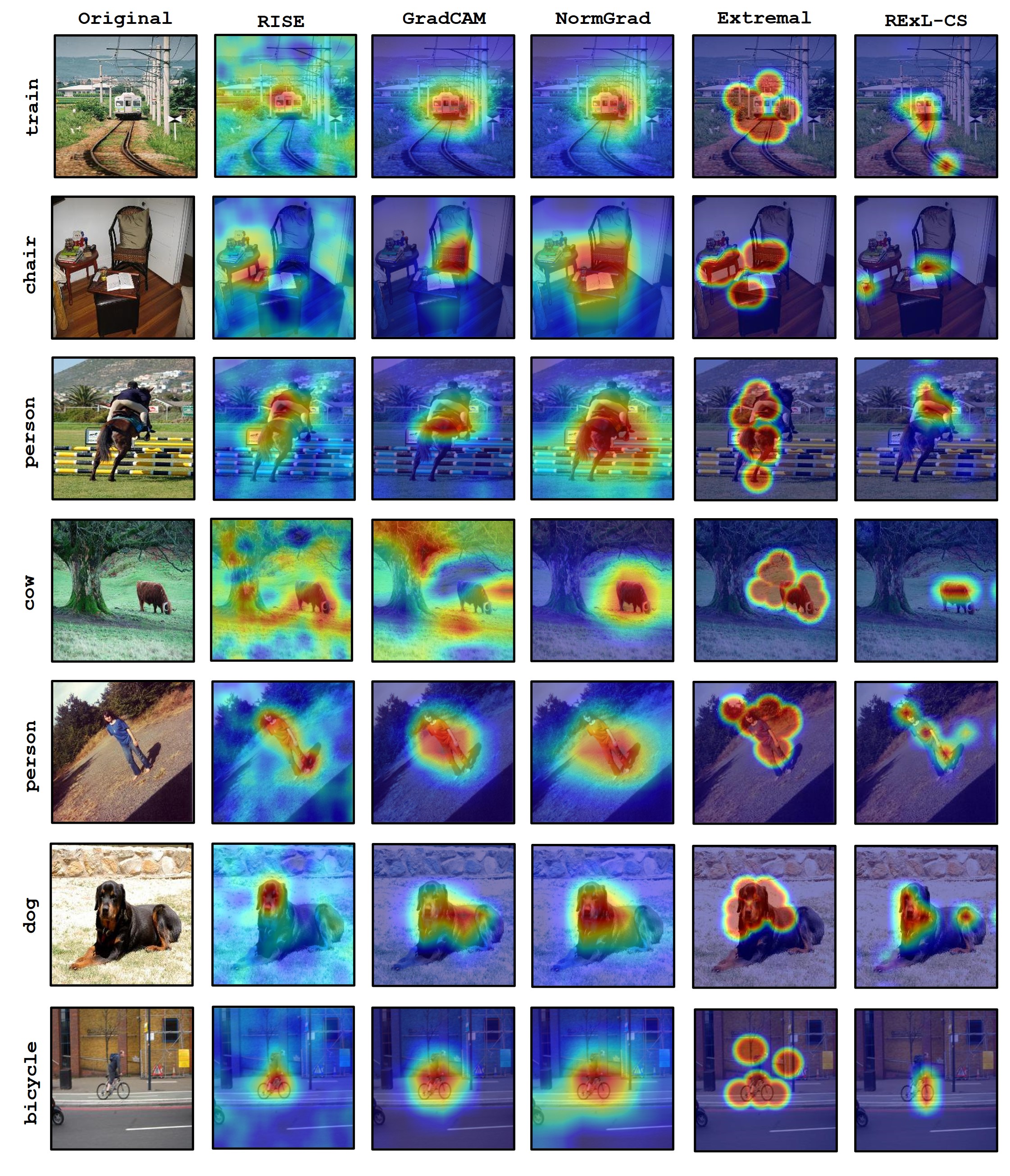}
    \caption{\small \textbf{Qualitative comparison of saliency maps for RExL-IS using ResNet$50$ base model on VOC$2007$:}For all the images, RExL-IS is significantly better than all the four baselines. While Extremal Perturbation provides almost equal importance to the entire object, white box methods (GradCAM and NormGrad) also simple add a large region of importance on the image instead of looking for the specific features. This is clearly highlighted in the "bicycle" image where GradCAM and NormGrad give large importance to the entire region containing the human as well.}
    \label{fig:single_img}
\end{figure}

These agents are trained on single images and explain the decisions made by the base model on that image only. These are hence not scalable for any practical datasets but if the base model needs to be explained very accurately and for a small number of images, RExL-IS will provide the best results. A comparison between the different baselines and RExL-IS can be seen in Figure (\ref{fig:single_img}). These images are taken from VOC$2007$ test split.

\newpage

\end{document}